\pdfoutput=1

\documentclass[11pt]{article}

\usepackage[]{acl}
\usepackage{times}
\usepackage{latexsym}

\usepackage[T1]{fontenc}

\usepackage[utf8]{inputenc}

\usepackage{microtype}

\usepackage{inconsolata}

\usepackage{graphicx}
\usepackage{multirow}
\usepackage{booktabs} 
\usepackage{amsmath}
\usepackage{hyperref}
\usepackage{listings} 
\usepackage{xcolor}
\usepackage{enumitem}

\definecolor{codeblue}{rgb}{0,0,,0.8}
\lstdefinestyle{mystyle}{
    keywordstyle=\color{codeblue},
    basicstyle=\ttfamily\footnotesize,
    breakatwhitespace=false,         
    breaklines=true,                 
    captionpos=b,                    
    keepspaces=true,                 
    numbers=left,                    
    numbersep=5pt,                  
    showspaces=false,                
    showstringspaces=false,
    showtabs=false,                  
    tabsize=2
}

\hypersetup{
    hidelinks= true,
    colorlinks = false,
    linkbordercolor = {white},
}

\title{AugCSE: Contrastive Sentence Embedding with Diverse Augmentations}

\author{Zilu Tang \\
  Boston University \\
  \texttt{zilutang@bu.edu} \\\And
  Muhammed Yusuf Kocyigit \\
  Boston University \\
  \texttt{kocyigit@bu.edu} \\\And
  Derry Wijaya \\
  Boston University \\
  \texttt{wijaya@bu.edu} \\}

\begin{document}
\maketitle
\begin{abstract}
Data augmentation techniques have been proven useful in many applications in NLP fields. Most augmentations are task-specific, and cannot be used as a general-purpose tool. In our work, we present AugCSE, a unified framework to utilize diverse sets of data augmentations to achieve a better, general purpose, sentence embedding model. Building upon the latest sentence embedding models, our approach uses a simple antagonistic discriminator that differentiates the augmentation types. With the finetuning objective borrowed from domain adaptation, we show that diverse augmentations, which often lead to conflicting contrastive signals, can be tamed to produce a better and more robust sentence representation. Our methods\footnote{Our code and data can be found at https://github.com/PootieT/AugCSE} achieve state-of-the-art results on downstream transfer tasks and perform competitively on semantic textual similarity tasks, using only unsupervised data.
\end{abstract}

\section{Introduction}

\begin{figure*}
\begin{center}
  \includegraphics[width=0.8\textwidth]{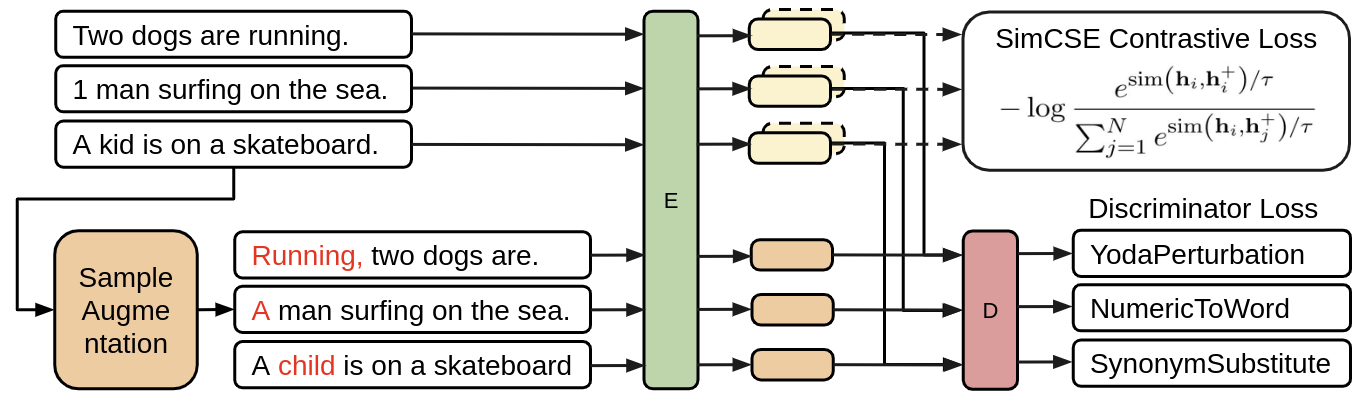}
  \caption{Overall framework of AugCSE. During training, each input sentence is randomly augmented with one of many augmentation methods. In addition contrastive loss from SimCSE, we add an antagonistic discriminator to predict the augmentation performed on the input example.
  }
  \label{fig:architecture}
  \vspace{-1.5em}
 \end{center}
\end{figure*}

Data augmentation in NLP can be useful in many situations, from low resource data setting, domain adaptation \cite{wei2021few}, debiasing \cite{dinan2020queens}, to improving generalization, robustness \cite{dhole2021nl}. In the vision domain, \citet{chen2020simple} shows that a diverse set of augmentation can be used to learn a robust general-purpose representation with contrastive learning. Similar work in sentence embedding space (\citealt{gao2021simcse, chuang2022diffcse}) has shown that a simple single augmentation such as dropouts from transformers \cite{devlin2019bert} can be used for contrastive objective. However, no previous work has thoroughly explored the impacts of a diverse set of augmentations with contrastive learning in the sentence embedding space. It is not straightforward to find the best augmentations that work for contrastive learning in different datasets or tasks \cite{gao2021simcse}. Single augmentation can instill invariance in models for a specific aspects of linguistic variability, while naively combining a diverse set of augmentations can lead to contradicting gradients, preventing models from generalizing well (Table~\ref{tab:ablation-discriminator})\footnote{Diverse augmentations have been shown to work without discriminator in vision \citep{chen2020simple}. We believe the difference resides in a much more structural distribution in natural language in comparison to images.}. In this work, we present AugCSE (Figure~\ref{fig:architecture}), a general approach to select and unify a diverse set of augmentations for the purpose of building a general-purpose sentence embedding. During training, in addition to using contrastive loss, we randomly perturb sentences with different augmentations and use a discriminator loss to unify embeddings from diverse augmentations. In short, our work presents the following key contributions:


\begin{itemize}[noitemsep,topsep=0pt]
  \item We show simple data augmentation methods can be used to improve individual tasks, while degrading performance on other tasks (due to shifted domain distribution).
  \item We present our simple discriminator objective that achieves competitive results on sentence similarity task (STS) and transfer classification tasks against state-of-the-art methods.
  \item We demonstrate through ablation and visualization that our model can unify contrasting distribution from diverse augmentations and that simple rule-based augmentations are sufficient for achieving competitive results.
\end{itemize}

\section{Background and Related Work}

\subsection{Contrastive learning}

Contrastive learning is shown to provide a clear signal to improve the embedding space, which is crucial for downstream tasks. 
The goal of contrastive learning is to use similar or dis-similar datapoints to regularize the embedding representation, such that similar datapoints (by human, or pre-defined standards) are embedded closer than those datapoints that aren't similar. Recently, many works in vision use contrastive objectives to obtain SOTA performance on image tasks from classification, detection, to segmentation using ImageNet (\citealp{deng2009imagenet, caron2018deep, chen2020simple, he2020momentum, caron2020unsupervised, grill2020bootstrap, zbontar2021barlow, chen2021exploring, bardes2022vicreg}). Most similar to our work is SimCLR \citep{chen2020simple}, which uses a diverse set of augmentation as positive contrastive pairs. In SimCLR, however, the procedure to obtain the best performing augmentation distribution was not clearly documented. Further, no previous work has investigated whether such an idea would work in the language domain. Our work provides a parallel investigation in NLP, accessing the usefulness of diverse augmentations in improving sentence representations. We also propose methodical procedures and heuristics on how such set of augmentations can be obtained given an end task.

\subsection{Sentence Embedding}
Building a general purpose sentence embedding model is useful for many tasks (\citealp{wang2021tsdae, izacard2021towards, gao2021condenser, gao2021simcse, chuang2022diffcse,chang2021deep}). SBERT \cite{reimers2019sentence} pioneered the efforts to improve semantic similarities between sentence embeddings using a siamese network with BERT \cite{devlin2019bert}. Finetuned with the natural language inference (NLI) dataset (\citealp{williams2018broad, bowman2015large}), SBERT predicts whether a hypothesis sentence entails or contradicts the second sentence. To tackle anisotropicness of BERT embedding space \cite{ethayarajh2019contextual}, \citet{li2020sentence} and \citet{su2021whitening} learn projection layer which converts BERT embedding to a Gaussian or zero-mean fixed-variance space. Following contrastive learning literature in vision, few works investigate alternative positive and negatives: from using different layers \cite{zhang2020unsupervised}, different models \cite{carlsson2020semantic}, against frozen model \cite{carlsson2020semantic}, different parts of document \cite{giorgi2021declutr}, to next sentences \cite{neelakantan2022text}.

With simplicity in mind, unsupervised SimCSE \cite{gao2021simcse} uses the same sentence with independent dropouts from transformers as positives and the rest of in-batch sentences as negatives, while supervised SimCSE uses NLI entailment sentence as positives, and contradiction as negatives. Lastly, the state-of-the-art method, DiffCSE \cite{chuang2022diffcse}, proposes to add an additional discriminative loss similar to ones used in ELECTRA \cite{clark2019electra}: the replaced token detection (RTD) loss to additionally increase the performance. The discriminator uses the original sentence embedding and a contextually perturbed sentence embedding to predict the token locations in which the two sentences differ. In contrast to DiffCSE, our discriminator predicts the augmentation type, a higher level task than predicting individual tokens. Additionally, our discriminator is in an antagonistic/adversarial relationship to our model, whereas the ELECTRA-like RTD objective is collaborative in nature. 

\subsection{NLP Augmentations}

 NLP augmentations are in more or less three flavors. Rule-based augmentations range from randomly deleting words, swap word orders \cite{wei2019eda}, to more structurally-sounds, or semantically specific ones (\citealp{zhang2015character}; \citealp{logeswaran2018content}). These simple augmentations, however, have been found to be not particularly effective in higher resource domain for task-agnostic purposes (\citealp{longpre2020effective}; \citealp{gao2021simcse}). The second kind of augmentations use pretrained language models (LM), to generate semantically similar examples. This area of work includes, but is not limited to back-translation (\citealp{li2019improving, sugiyama2019data}), paraphrase models (\citealp{li2019decomposable, li2018paraphrase, iyyer2018adversarial}), style transfer models (\citealp{fu2018style, krishna2020reformulating}), contextually perturbed models (\citealp{morris2020textattack, jin2020bert}), to large LM-base augmentation (\citealp{kumar2020data}; \citealp{yoo2021gpt3mix}). Lastly, a few methods generate augmentations in the embedding space. These methods often perform interpolation (\citealp{devries2017dataset,chen2020mixtext}), noising \cite{kurata2016labeled}, and autoencoding (\citealp{schwartz2018delta, kumar2019closer}) with embedded data points. However, due to the discreteness of NL \cite{bowman2016generating} and anisotropy \cite{ethayarajh2019contextual}, the introduced noise often outweighs the benefit of additional data.

Recently, NL-Augmenter \cite{dhole2021nl} collected over 100 augmentation methods, with the intention to provide robustness diagnostics for NLP models against different type of data perturbations\footnote{https://github.com/GEM-benchmark/NL-Augmenter}. In our work, we show that a diverse set of augmentations, even with simple rule-based augmentations, which are cheaper and  more controllable than LM-based augmentations, can be used to learn robust general-purpose sentence embedding.


\section{Motivation}

\subsection{Single augmentation is task specific}
Augmentations, especially ones that exploit surface level semantics using simple rules, are task specific and have been used alone only if the augmentation aligns with the task objective for the dataset \cite{longpre2020effective}. For instance, \citet{dinan2020queens} changes gendered words in a sentence to instill gender invariance for bias mitigation. Inspired by hard negative augmentations in contrastive learning (\citealp{gao2021simcse, sinha2020negative}), we use the following case studies to reinforce the conclusion from the perspective of negative data augmentation. In both scenarios, we use the negative augmentations ($\textbf{h}_i^-$) loss (with positive examples $\textbf{h}_i^+$) for contrastive objective \cite{gao2021simcse}:
\vspace{-1.0em}
\label{eqn: contrastive-neg}
\begin{align}
-\mathrm{log} \frac{e^{\mathrm{sim}(\textbf{h}_i, \textbf{h}_i^+)/\tau}}{\sum_{j=1}^N (e^{\mathrm{sim}(\textbf{h}_i, \textbf{h}_i^+)/\tau} + e^{\mathrm{sim}(\textbf{h}_i, \textbf{h}_i^-)/\tau})}
\vspace{-0.7em}
\end{align}

\noindent where $\mathrm{sim}$ is cosine similarity, $\tau$ is the temperature parameter controlling for the contrastive strength, and $N$ is batch size. Since some augmentations do not have 100\% perturbation rate, we remove  datapoints that do not have a successful negative augmentation. For the remaining datapoints, we use original sentences as positives, and train with different augmentations as the negatives. In addition, we also present average transfer tasks \cite{conneau2018senteval} performance as a metric for embedding quality (\textbf{trans.}, detailed in Sec \ref{mainexp}).

\paragraph{Case study 1: linguistic acceptability} We first test embedding performance on CoLA \cite{warstadt2018neural}, a binary sentence classification task predicting linguistically acceptability. If an augmentation frequently introduces grammatical errors, it should perform well as a negative. 

\paragraph{Case study 2: contradiction vs. entailment} Natural language inference (NLI) datasets (\citealp{bowman2015large,williams2018broad}) provide triplets of sentences: an hypothesis, a sentence entailing, and a sentence in contradiction to the hypothesis. A good embedding should place the entailment sentence closer to the hypothesis than the contradiction sentence, and in fact, that is the exact hypothesis exploited by supervised SimCSE. We calculate the similarity between hypothesis and an entailment sentence and similarity between hypothesis and a contradiction sentence, and count how often is the former larger than the later in ANLI \cite{nie2020adversarial}. If an augmentation can reverse the semantics of sentences, then it should perform well as a negative.

\begin{table}[t]
    \centering
    \small
    \begin{tabular}{p{0.65\linewidth} p{0.10\linewidth} p{0.10\linewidth}}
    \toprule
        Augmentation   & \textbf{CoLA}  & \textbf{trans.}\\  \midrule
        BERT\textsubscript{base}  & 75.93 & 84.66 \\
        Unsupervised SimCSE\textsubscript{BERT} & 71.91 & \textbf{85.81} \\  
        RandomContextualWordAugmentation & \textbf{78.14} & 80.51\\ 
        SentenceSubjectObjectSwitch & 76.80 & 80.31 \\\midrule

        Augmentation   &  \textbf{ANLI} & \textbf{trans.} \\  \midrule
        BERT\textsubscript{base}  & 53.80 & 84.66  \\ 
        Unsupervised SimCSE\textsubscript{BERT} & 53.42 & \textbf{85.81} \\
        AntonymSubstitute & \textbf{58.78} & 79.93 \\ 
        SentenceAdjectivesAntonymsSwitch & 58.63 & 80.11 \\ \bottomrule
    \end{tabular}
    \caption{ Top negative augmentations for CoLA and ANLI, both measured in accuracy, with average transfer performance. See augmentation description in \ref{apx: aug-desc}
    \label{tab:one-neg-aug}}
    \vspace{-2em}
\end{table}

\paragraph{Insight:} As expected (Table~\ref{tab:one-neg-aug}), augmentations known to introduce a lot of grammatical mistakes: \hyperref[apx: RandomContextualWordAugmentation]{RandomContextualWordAugmentation} \cite{zang2020word} performs the best in \textbf{CoLA} and those that reverse semantics: \hyperref[apx: AntonymSubstitute]{AntonymSubstitute}, and \hyperref[apx: SentenceAdjectivesAntonymsSwitch]{SentenceAdjectivesAntonymsSwitch} performs well in \textbf{ANLI}. However, single augmentation significantly under-performs in \textbf{trans}fer tasks, reducing robustness. This suggests the need for diverse augmentations (\citealp{chen2020simple,ren2021text}). 


\subsection{Difficulty of selecting contrastive pairs}

\citet{gao2021simcse} experimented with a combination of MNLI \citep{williams2018broad} and SNLI \citep{bowman2015large} and found that using entailment as positives and contradictions as negatives performs well. In addition to this setting, we performed additional ablations to show that it is usually unclear which sentence pair dataset or augmentation would provide the best result as contrastive pairs (Table \ref{tab:ablation-simcse-nli}). Sometimes, non-intuitive pairs could yield decent results\footnote{See more discussion on negation in deep learning in ~\ref{apx: negation}}. Together with the specificity of individual augmentations, this motivates for a general framework to select and combine multiple augmentations to achieve a robust, general-purpose embedding.

\begin{table}[t]
    \centering
    \small
    \begin{tabular}{p{0.79\linewidth} p{0.11\linewidth}}
    \toprule
        Trial                      & \textbf{STS-b} \\ \midrule
        unsupervised SimCSE        & 81.18 \\
        supervised SimCSE          & 85.64 \\ 
        no contradiction           & 83.60 \\
        contradiction as pos & 79.55 \\
        contradiction as pos, entailment as neg & 67.16  \\ 
        supervised SimCSE w/ ANLI  & 75.99 \\\bottomrule
    \end{tabular}
    \caption{ Alternative choices of positives and negatives with SimCSE. All results are reproduced by us.
    \label{tab:ablation-simcse-nli}}
    \vspace{-1.5em}
\end{table}

\section{Methods}

\subsection{Augmentation Selection}
\citet{dhole2021nl} introduced 100+ augmentation methods. We also added non-duplicating augmentation methods from popular repositories: nlpaug, checklist, TextAugment, TextAttack, and TextAutoAugment (\citealt{ma2019nlpaug, ribeiro2020beyond, marivate2020improving, morris2020textattack, ren2021text}), including \hyperref[apx: RandomDeletion]{RandomDeletion}, \hyperref[apx: RandomSwap]{RandomSwap}, \hyperref[apx: RandomCrop]{RandomCrop}, \hyperref[apx: RandomWordAugmentation]{RandomWordAugmentation}, \hyperref[apx: RandomWordEmbAugmentation]{RandomWordEmbAugmentation}, and \hyperref[apx: RandomContextualWordAugmentation]{RandomContextualWordAugmentation}\footnote{SimCSE tried RandomDeletion, RandomCrop; DiffCSE tried RandomDeletion, RandomInsertion, and their RTD is based on RandomContextualWordAugmentation.}.

To narrow down the augmentations we experiment with, we selected for single-sentence augmentations that are either labeled \textbf{highly meaning preserving}, \textbf{possible meaning alteration}, or \textbf{meaning alteration}. After preliminary filtering (Appendix~\ref{apx: narrow-down}), Table~\ref{tab: aug-list} contains all augmentations we included in our experiments. To select for a diverse set of augmentation for main results in STS-b and transfer tasks, we trained models using single augmentation as positives, and pick augmentations that obtained top performance on STS-B and transfer tasks. For full single augmentation results see Appendix~\ref{apx: single-aug}.



\begin{table*}[ht]
    \centering
    \small
    \begin{tabular}{p{0.3\linewidth}p{0.35\linewidth}p{0.25\linewidth}}
    \toprule
    Meaning Alteration & Possible Meaning Alteration & Highly Meaning Preserving\\ \midrule 
      \textbf{\underline{\hyperref[apx: SentenceAdjectivesAntonymsSwitch]{SentenceAdjectivesAntonymsSwitch}}}, \textbf{\underline{\textcolor{orange}{\hyperref[apx: SentenceAuxiliaryNegationRemoval]{SentenceAuxiliaryNegationRemoval}}}}, \hyperref[apx: hypernym/hyponym]{ReplaceHypernyms}, \hyperref[apx: hypernym/hyponym]{ReplaceHyponyms}, \textbf{\underline{\textcolor{orange}{\hyperref[apx: SentenceSubjectObjectSwitch]{SentenceSubjectObjectSwitch}}}}, \textbf{\underline{\hyperref[apx: CityNamesTransformation]{CityNamesTransformation}}} \hyperref[apx: AntonymSubstitute]{AntonymSubstitute} &

       \textbf{\hyperref[apx: ColorTransformation]{ColorTransformation}},\hyperref[apx: Summarization]{Summarization}, \textbf{\underline{\textcolor{orange}{\hyperref[apx: DiverseParaphrase]{DiverseParaphrase}*}}},\textbf{\underline{\hyperref[apx: SentenceReordering]{SentenceReordering}}}, \textbf{\underline{\textcolor{orange}{\hyperref[apx: TenseTransformation]{TenseTransformation}*}}},\hyperref[apx: RandomDeletion]{RandomDeletion}, \hyperref[apx: RandomCrop]{RandomCrop}, \textbf{\underline{\textcolor{orange}{\hyperref[apx: RandomSwap]{RandomSwap}*}}}, \textbf{\hyperref[apx: RandomWordAugmentation]{RandomWordAugmentation}}, \hyperref[apx: RandomWordEmbAugmentation]{RandomWordEmbAugmentation}, \hyperref[apx: RandomContextualWordAugmentation]{RandomContextualWordAugmentation} &

      \textbf{\underline{\hyperref[apx: YodaPerturbation]{YodaPerturbation}}}, \textbf{\underline{\textcolor{orange}{\hyperref[apx: ContractionExpansions]{ContractionExpansions}*}}}, \textbf{\underline{\textcolor{orange}{\hyperref[apx: DiscourseMarkerSubstitution]{DiscourseMarkerSubstitution}}}}, \textbf{\underline{\textcolor{orange}{\hyperref[apx: Casual2Formal]{Casual2Formal}}}}, \textbf{\hyperref[apx: GenderSwap]{GenderSwap}}, \hyperref[apx: GeoNamesTransformation]{GeoNamesTransformation}, \textbf{\hyperref[apx: NumericToWord]{NumericToWord}}, \hyperref[apx: SynonymSubstitution]{SynonymSubstitution} \\
     \bottomrule
    \end{tabular}

    \caption{Final subsets of augmentations included in experiments. Augmentations in 16-Aug experiments are \textbf{bolded}, 12-Aug experiments are \underline{underlined}, 8-Aug experiments are colored \textcolor{orange}{orange} and 4-Aug experiments marked with asterisks(*). For full descriptions of augmentations, see Appendix \ref{apx: aug-desc}.
    \label{tab: aug-list}}
    \vspace{-1.5em}
\end{table*}

\subsection{Augmentation Sampling}
 To save computation and control for randomness, we augment the training dataset once for every augmentation and cache the results. Prior to training, augmentations are read from caches and uniformly sampled at each data point. Since not every augmentation perturbs the original sentence at every data point, we then correct augmentation label to "no augmentation" if the augmented sentence is the same as original sentence. This leads to a larger portion of the sentence having the label "no augmentation" than each individual augmentation\footnote{We also tried resampling augmentations between each epochs and found that to underperform fixed sampling.}.

\subsection{Model Architecture}
In our experiments, we train sentence embedding encoders using BERT- and RoBERTa-base for fair comparison to previous methods: SimCSE and DiffCSE. 
During training, we pass sentence representations through 2-layer projection layer with batchnorm, introduced by DiffCSE. We remove projection layers during inference and obtain sentence embeddings directly from the encoder. Formally, we train with contrastive loss, shown in the equation at the top right of Figure~\ref{fig:architecture}. We refer to this contrastive loss as $\mathcal{L}_{contrastive}$. 
We use the embedding corresponding to \textbf{[CLS]} token as sentence embedding in all experiments.

Contrastive loss regularizes on individual data pair level, which is a very strict constraint to resolve distributional shifts that augmentations introduce. To train sentence encoders that are invariant with respect to the shifts between diverse augmentations, we introduce an antagonistic discriminator. We pass the concatenated embeddings of original and augmented sentences into the discriminator (code in Appendix~\ref{apx: mlp-code}) trained with the  $\mathcal{L}_{discriminator}$ loss, defined as binary cross entropy between predicted and actual augmentations:

\begingroup
\setlength\abovedisplayskip{0pt}
\small
\begin{gather}
     -\frac{1}{K}\sum_{i=1}^K y_i \mathrm{log}(p(y_i)) + (1-y_i) \mathrm{log}(1-p(y_i))
\end{gather}
\setlength\belowdisplayskip{0pt}

\endgroup
\noindent where $K$ is the number of augmentation types (plus "no augmentation"), and $p(y_i)$ is the probability of augmentation type $i$ predicted by the discriminator. To encourage  augmentation-invariant encoder, the first layer of the discriminator uses a gradient reversal layer (\citealt{ganin2015unsupervised, zhu2015aligning, ganin2016domain}) (code in Appendix~\ref{apx: grl-code}) that allows the gradient to be multiplied with a negative multiplier $\alpha$ in backward pass such that while discriminator is trained to minimize discriminator loss, the encoder is trained to maximize the discriminator loss all in one pass. We find this simple scheme to work well without having to deal with the instability around training adversarial networks (\citealt{creswell2018generative, clark2019electra}). 

Finally, the overall loss of our model (AugCSE):
\vspace{-0.5em}
\begin{equation}
    \mathcal{L} = \mathcal{L}_{contrastive} + \lambda * \mathcal{L}_{discriminator}
\end{equation}
where $\lambda$ is a coefficient that tunes the strength of discriminator loss.

\section{Experiments}
\label{mainexp}
\subsection{Evaluation Datasets}
For fair comparison, we use the same dataset SimCSE used: 1M sentences randomly selected from Wikipedia. After training, we use frozen embeddings to evaluate our method on 7 semantic textual similarity (STS) tasks and 7 (SentEval) transfer tasks \cite{conneau2018senteval}. STS tasks include \textbf{STS 2012 - 2016} \cite{agirre2016semeval}, \textbf{STS-Benchmark} \cite{cer2017semeval}, and \textbf{SICK-Relatedness} \cite{marelli2014sick}. In STS tasks, Spearman correlation is calculated between model's embedding similarity of the pair of sentences against human ratings (1-5). Transfer tasks are single sentence classification tasks from SentEval including \textbf{MR} \cite{pang2005seeing}, \textbf{CR} \cite{hu2004mining}, \textbf{MPQA} \cite{wiebe2005annotating}, \textbf{MRPC} \cite{dolan2005automatically}, \textbf{TREC} \cite{voorhees2000building}, \textbf{SST-2} \cite{socher2013recursive}, and \textbf{SUBJ} \cite{pang2004sentimental}. We follow the standard evaluation setup from \cite{conneau2018senteval}, training a logistic regression classifier on top of frozen sentence embeddings. See Appendix~\ref{apx: hyp-selection} for details on hyperparameter search.

\subsection{Evaluation Baselines}
We include several levels of baselines. From word-averaged Glove embedding \cite{pennington2014glove}, to BERT\textsubscript{base}, using both average pooling as well as [CLS] token. We include post processing methods, \textbf{BERT-flow} \cite{li2020sentence}, and \textbf{BERT-whitening} \cite{su2021whitening}, as well as other more recent contrastive sentence embeddings: \textbf{CT-BERT} \cite{carlsson2020semantic}, \textbf{SG-OPT} \cite{kim2021self}, \textbf{SimCSE} \cite{gao2021simcse}, \textbf{DiffCSE} \cite{chuang2022diffcse}. We also report results from \textbf{DeCLUTER} \cite{giorgi2021declutr} and \cite{neelakantan2022text} (\textbf{cpt-text-S}) as a comparison for what larger model and larger training data size would benefit. More specifically, DeCLUTER mines positives from documents, and cpt-text-S uses next sentence as positives.

\subsection{STS Results}

\begin{table*}[t]
    \centering
    \small
    \begin{tabular}{lcccccccc}
    \toprule
    Model                               & STS12 & STS13 & STS14 & STS15 & STS16 & STS-B & SICK-R & Avg.\\ \midrule
    GloVe embeddings (avg.) $\clubsuit$        & 55.14 & 70.66 & 59.73 & 68.25 & 63.66 & 58.02 & 53.76 & 61.32\\ 
    BERT\textsubscript{base} (first-last avg.) $\diamondsuit$    & 39.70 & 59.38 & 49.67 & 66.03 & 66.19 & 53.87 & 62.06 & 56.70\\ 
    BERT\textsubscript{base}-flow $\diamondsuit$                 & 58.40 & 67.10 & 60.85 & 75.16 & 71.22 & 68.66 & 64.47 & 66.55\\ 
    BERT\textsubscript{base}-whitening $\diamondsuit$            & 57.83 & 66.90 & 60.90 & 75.08 & 71.31 & 68.24 & 63.73 & 66.28\\ 
    SG-OPT-BERT\textsubscript{base} †              & 66.84 & 80.13 & 71.23 & 81.56 & 77.17 & 77.23 & 68.16 & 74.62\\ 
    Unsupervised SimCSE-BERT\textsubscript{base} $\diamondsuit$  & 68.40 & 82.41 & 74.38 & 80.91 & 78.56 & 76.85 & \textbf{72.23} & 76.25\\ 
    DiffCSE-BERT\textsubscript{base} $\heartsuit$    & \textbf{72.28} & \textbf{84.43} & \textbf{76.47} & \textbf{83.90} & \textbf{80.54} & \textbf{80.59} & 71.23 & \textbf{78.49}\\ \midrule
    * AugCSE-BERT\textsubscript{base}              & \underline{71.40} & \underline{83.93} & \underline{75.59} & \underline{83.59} & \underline{79.61} & \underline{79.61} & \underline{72.19} & \underline{77.98} \\ \midrule
    
    RoBERTa\textsubscript{base} (first-last avg.) $\diamondsuit$ & 40.88 & 58.74 & 49.07 & 65.63 & 61.48 & 58.55 & 61.63 & 56.57\\ 
    RoBERTa\textsubscript{base}-whitening $\diamondsuit$         & 46.99 & 63.24 & 57.23 & 71.36 & 68.99 & 61.36 & 62.91 & 61.73\\ 
    Unsupervised SimCSE-RoBERTa\textsubscript{base} $\diamondsuit$           & \textbf{70.16} & 81.77 & 73.24 & 81.36 & 80.65 & 80.22 & 68.56 & 76.57\\ 
    DiffCSE-RoBERTa\textsubscript{base} $\heartsuit$    & \underline{70.05} & \textbf{83.43} & \textbf{75.49} & \textbf{82.81} & \textbf{82.12} & \textbf{82.38} & \textbf{71.19} & \textbf{78.21}\\ \midrule
    * AugCSE-RoBERTa\textsubscript{base}              & 69.30 & \underline{82.17} & \underline{73.49} & \underline{81.82} & \underline{81.40} & \underline{80.86} & \underline{68.77} & \underline{76.83} \\ \bottomrule
                              \multicolumn{9}{c}{Larger Training Data / Model Size} \\ \midrule
    DeCLUTR-RoBERTa\textsubscript{base} $\diamondsuit$          & 52.41 & 75.19 & 65.52 & 77.12 & 78.63 & 72.41 & 68.62 & 69.99\\ 
    CPT-text-S $\spadesuit$                 & 62.1 & 60.0 & 62.0 & 71.8 & 73.7 & - & - & - \\
    \bottomrule
    \end{tabular}
    \caption{STS Test Set Performance (Spearman's correlation) from different sentence embedding models. $\clubsuit$: results from \citep{reimers2019sentence}. $\diamondsuit$: results from \citep{gao2021simcse}. †: results from \citep{kim2021self}. $\heartsuit$: results from \citep{chuang2022diffcse}. Best results are \textbf{bolded}, second best results are \underline{underlined} 
    \label{tab:sts-test}}
    \vspace{-1em}
\end{table*}

We show STS test results in Table~\ref{tab:sts-test}. AugCSE performs competitively against SOTA methods, with both BERT and RoBERTa. AugCSE also outperforms larger models trained with more data (DeCLUTR and cpt-text-s). We discuss this in Sec~\ref{sec: lim-sts}.

\subsection{Transfer Tasks Results}

\begin{table*}[t]
    \centering
    \small
    \begin{tabular}{lcccccccc}
    \toprule
    Model                                  & MR    & CR    & SUBJ  & MPQA  & SST   & TREC  & MRPC  & Avg. \\ \midrule
    
    GloVe embeddings (avg.) $\clubsuit$               & 77.25 & 78.30 & 91.17 & 87.85 & 80.18 & 83.00 & 72.87 & 81.52 \\
    Avg. BERT embeddings $\clubsuit$                  & 78.66 & 86.25 & 94.37 & 88.66 & 84.40 & \textbf{92.80} & 69.54 & 84.94 \\
    BERT-[CLS]embedding $\clubsuit$                   & 78.68 & 84.85 & 94.21 & 88.23 & 84.13 & \underline{91.40} & 71.13 & 84.66 \\
    SimCSE-BERT\textsubscript{base} $\diamondsuit$        & 81.18 & 86.46 & 94.45 & 88.88 & 85.50 & 89.80 & 74.43 & 85.81 \\
     w/ MLM                                 & \underline{82.92} & \underline{87.23} & \textbf{95.71} & 88.73 & \underline{86.81} & 87.01 & \textbf{78.07} & 86.64  \\
    DiffCSE-BERT\textsubscript{base} $\heartsuit$  & 82.69 & \underline{87.23} & 95.23 & \underline{89.28} & 86.60 & 90.40 & \underline{76.58} & \underline{86.86} \\\midrule
    * AugCSE-BERT\textsubscript{base}      & \textbf{82.88} & \textbf{88.19} & \underline{95.40} &\textbf{89.43} & \textbf{87.15} & \underline{91.40} & 75.07 & \textbf{87.07} \\ \midrule
    
    SimCSE-RoBERTa\textsubscript{base} $\diamondsuit$     & 81.04 & 87.74 & 93.28 & 86.94 & 86.60 & 84.60 & 73.68 & 84.84 \\
     w/ MLM                                 & \textbf{83.37} & 87.76 & \textbf{95.05} & 87.16 & \textbf{89.02} & \textbf{90.80} & 75.13 & \underline{86.90} \\
    DiffCSE-RoBERTa\textsubscript{base} $\heartsuit$   & \underline{82.82} & \textbf{88.61} & \underline{94.32} & \textbf{87.71} & \underline{88.63} & \underline{90.40} & \textbf{76.81} & \textbf{87.04} \\\midrule
    * AugCSE-RoBERTa\textsubscript{base}     & \underline{82.82} & \underline{88.48} & 93.72 & \underline{87.40} & 86.82 & 88.80 & \underline{75.88} & 86.27 \\ \midrule
    
                          \multicolumn{9}{c}{Larger Training Data / Model Size} \\ \midrule
    DeCLUTR-RoBERTa\textsubscript{base} † & 85.16 & 90.68 & 95.78 & 88.52 & 90.01 & 93.20 & 74.61 & 88.28  \\
    CPT-text-S $\spadesuit$                 & 87.1  & 90.1 & 94.9 & 88.3 & 91.8 & 95.2 & 71.6 & 88.4 \\
    \bottomrule

    \end{tabular}
    \caption{SentEval Test Set Performance (accuracy) from different sentence embedding models. $\clubsuit$: results from \citep{reimers2019sentence}. $\diamondsuit$: results from \citep{gao2021simcse}. †: results from \citep{giorgi2021declutr}. $\heartsuit$: results from \citep{chuang2022diffcse}. DeCLUTR was finetuned on ~500K documents $\spadesuit$: results from \citep{neelakantan2022text}. CPT-text-S models has 300M parameters and is trained on "Internet data".
    \label{tab:transfer-test}}
    \vspace{-1.5em}
\end{table*}

We show transfer tasks test set results in Table ~\ref{tab:transfer-test}. With BERT\textsubscript{base} 
AugCSE outperforms DiffCSE in average transfer score and improve 4 out of 7 SentEval tasks. In RoBERTa\textsubscript{base}, we still see competitive performance. Here, larger models with more training data outperform existing methods. 

\subsection{Discriminator Objective Variations}


\begin{table}[t]
    \centering
    \small
    \begin{tabular}{ccc}
    \toprule
        discriminator    & STS-b         & Transfer\\ \midrule
        AugCSE           & \textbf{85.25}& \textbf{85.80}   \\
        bool             & 84.52         & 85.44   \\ 
        positive         & 84.54         & 85.78   \\ 
        no discriminator & 84.91         & 85.25   \\\bottomrule
    \end{tabular}
    \caption{ Dev performance varying discriminator types.
    \label{tab:ablation-discriminator}}
    \vspace{-1em}
\end{table}

In addition to predicting the augmentation type (\textbf{AugCSE}), we vary the discriminative objectives in Table~\ref{tab:ablation-discriminator}. With \textbf{bool}, the discriminator predicts whether the second sentence is augmented or not (since not every augmentation is guaranteed 100\% perturbation rate). 
With \textbf{positive}, we use augmented sentence as positives in the contrastive loss as well as using their augmentation types in the discriminator loss. For this setting, we use a symmetric loss similar to one in CLIP \cite{radford2021learning} to boost performance because contrasting two different distributions from augmented and natural text benefits from a symmetric regularization. In \textbf{no discriminator}, we use augmented sentence as positives in the contrastive loss but do not use a discriminator, which is the most naive way of using augmentation in contrastive learning (as in SimCLR\cite{chen2020simple}). Empirically, we found that using augmentations only for the discriminative objective (\textbf{AugCSE}) performs the best and improves transfer results significantly over \textbf{no discriminator}. To understand such phenomenon, we can think of the discriminative objective as a weaker form of regularization, where we enforce invariance on the augmentation distribution level, rather than on individual augmented sentence level. The weaker constraint tolerates more noise in augmentation while distributionally improves the embedding space. Intuitively it make sense because the "noises" we introduce with augmentations do not impact the semantics of each sentence equally (e.g. randomly dropping an article in a sentence changes the semantics much less than dropping a verb). However, with the discriminative objective we do encourage that such noise be tolerated on a distributional level. This subtle difference is analogous to works in AI fairness, where antagonistic discriminator optimizes for group fairness \cite{chouldechova2020snapshot}, while contrastive learning optimizes for individual fairness \cite{dwork2012fairness}.


We also experiment with different values of the $\alpha$ in gradient reversal layer in Table~\ref{tab:ablation-alpha}. Since $\alpha$ is a constant multiplied to the gradient from the discriminator and applied to downstream encoder, changing $\alpha=-1$ to $\alpha=1$ is equivalent to changing discriminator from being antagonistic (AugCSE) to being collaborative (similar to DiffCSE). The magnitude determines how antagonistic or collaborative the discriminator is. We can see that the discriminator being antagonistic is crucial for our model performance (more detailed explorations and visualizations of the impact of $\alpha$ and on the embedding space are shown in Fig.~\ref{fig: alpha-emb} and \ref{fig: alpha-train} in the Appendix).

\begin{table}[t]
    \centering
    \small
    \begin{tabular}{ccc}
    \toprule
        $\alpha$  & STS-b & Transfer\\ \midrule
        100     & 60.47 & 85.68 \\
        10      & 72.33 & 85.67 \\
        1       & 80.85 & 85.78 \\ 
        -1 (AugCSE)      & \textbf{85.25} & \textbf{85.80} \\ 
        -10     & 84.68 & 85.68 \\
        -100    & 80.54 & 85.67 \\\bottomrule
    \end{tabular}
    \caption{ Dev performance with various $\alpha$ values.
    \label{tab:ablation-alpha}}
    \vspace{-1.5em}
\end{table}

\subsection{Augmentation ablation}

We also vary the number of augmentation to determine the importance of diversity of augmentation for performance. For improving STS performance, we found 8 augmentations (Table~\ref{tab:ablation-aug}) to be a sweet spot between including as diverse set of augmentations and keeping the augmentations relevant to the task. We see that including additional augmentation (16) can help further improve transfer results, but we use 8 augmentations in our main results for its simplicity. It is possible that we can improve our results further by including more diverse set of augmentations, we leave that for future studies.

\begin{table}[t]
    \centering
    \small
    \begin{tabular}{ccc}
    \toprule
        Trial   & STS-b  & Transfer \\ \midrule
        4-Aug   & 84.97  & 85.79    \\ 
        8-Aug   & \textbf{85.25}  & 85.80 \\ 
        12-Aug  & 84.63  & 85.73 \\
        16-Aug  & 84.83  & \textbf{85.92} \\ \bottomrule
    \end{tabular}
    \caption{Ablation varying augmentations size. 
    \label{tab:ablation-aug}}
    \vspace{-1em}
\end{table}

\subsection{Pretrained model based augmentation}

LM-enabled augmentations could, in theory, beat the combination of all other augmentations by generating a diverse set of paraphrases using linguistic priors from training data. 
In 8, 12, and 16 augmentation setting, only \hyperref[apx: DiverseParaphrase]{\textbf{DiverseParaphrase}} and \hyperref[apx: Casual2Formal]{\textbf{Casual2Formal}} augmentations use pretrained model. To see how crucial LM-based augmentations are to our performance,  we 
remove these augmentations and compare results with original settings. Without LM-based augmentations, we still see comparable results as before (Table~\ref{tab:ablation-lm-aug}). STS results actually \textbf{improve} across all trials. 

\begin{table}[t]
    \centering
    \small
    \begin{tabular}{ccc}
    \toprule
        Trial  &  STS-b ($\Delta$) & Transfer ($\Delta$)\\ \midrule
        8-2-Aug  & 85.31 (\textcolor{green}{+0.06})   & 85.74 (\textcolor{red}{-0.06})      \\ 
        12-2-Aug & 84.83 (\textcolor{green}{+0.20})   & 85.83 (\textcolor{green}{+0.10})  \\
        16-2-Aug & 84.84 (\textcolor{green}{+0.01})   & 85.78 (\textcolor{red}{-0.14})  \\ \bottomrule
    \end{tabular}
    \caption{ Performance after removing LM-based augmentations. Colored numbers indicate deltas compared to augmentation sets that include LM-based augs.
    \label{tab:ablation-lm-aug}}
    \vspace{-1.5em}
\end{table}





\section{Analysis and Discussion}

In our experiments, we selected subsets of top performing augmentations by looking at their  individual finetuned performances. Such selection procedure may not be feasible due to resource constraints. In the following sections and in App.~\ref{apx: isomorphism}, we discuss a few metrics that 
could be used to provide some signal in selecting the best augmentation (or dataset) for contrastive learning.  We also discuss the broader impact of our work, advantages, and yet unresolved problems in the field.

\subsection{Similarity and perplexity}
One simple way of measuring point-wise distance between original and augmented sentences is using semantic similarity (approximated with cosine similarity between their SBERT embeddings\footnote{sentence-transformers/all-mpnet-base-v2}) and perplexity difference (calculated with GPT2 \cite{sanh2019distilbert}). Across all augmentations, similarities have positive correlation with STS-b and Transfer performance (Pearson correlation coefficients of 0.72 and 0.6, resp.) while perplexities difference have negative correlation with STS-b and Transfer performance (coefficients of -0.53 and -0.58, resp.) when augmentations are used as positives. This indicates that augmented sentences with higher similarities and lower perplexities differences to the originals may be useful as positive examples in contrastive learning. 
For more results and correlation with other metrics such as embedding isomorphism, see Appendix~\ref{apx: isomorphism} and ~\ref{apx: single-aug}.


\subsection{Domain shift in augmentation}
In Figure~\ref{fig:pca} in the Appendix, we visualize the embedding distribution of sampled sentences pre- and post- augmentations, of pretrained BERT and AugCSE\textsubscript{BERT}. We observe that augmentations do introduce distributional shift and that our discriminator can indeed unify distributions from diverse augmentations, along with evidence that $\alpha$ also impact unification (Figure~\ref{fig: alpha-emb} in the Appendix).


\subsection{LM-based vs. rule-based augmentations}
In our experiments, we observe that our model (AugCSE) performance does not depend on LM-based augmentations. 
AugCSE performance matches that of DiffCSE (that uses solely LM-based augmentation) and in many cases, removing LM-based augmentations even improves its performance (Table~\ref{tab:ablation-lm-aug}). 
This is an added advantage given that LM-based augmentations may be more expensive to run, are not as controllable as rule-based augmentations, and may contain bias learned from text in the wild that can reinforce undesirable properties in the sentence embedding. In comparison, rule-based models can precisely control for such behaviors, mitigate bias \cite{dinan2020queens}, or introduce invariance in embedding space specific to the needs of the downstream tasks.

\section{Conclusion}
We present AugCSE, a general framework that combines diverse sets of augmentations to improve general sentence embeddings. In addition to the contrastive loss, we introduce an antagonistic discriminator that loosely constrain the model to become invariant to distributional shifts created from augmentations. In addition to outperforming previous methods, our framework is much more controllable, which has an added advantage of being able to mitigate undesirable properties from pretrained LMs, which inherit bias and toxicity from training data on the internet. Additionally, AugCSE can work with cheaper augmentations to run, resulting in a more resource-friendly approach to training generic sentence embedding models.

\section*{Limitations}

\paragraph{Semantic textual similarity for evaluation.}
\label{sec: lim-sts}
Sentence embedding literature has focused primarily on evaluating models using sentence semantic similarity tasks and SentEval transfer tasks. While transfer tasks may capture a wider range of desirable properties for a generic sentence embedding model, STS is often not a perfect indicator of sentence embedding quality. As noted by \citet{neelakantan2022text}, STS tasks performance decreases as transfer task performance increases. This trend can also be observed in other robust models such as DeCLUTR. In future studies, we urge users to use STS tasks as only a subset of the transfer tasks when evaluating sentence embedding.

However, sentence semantic is still an important and difficult task that is not yet solved especially when considering the recursive structure, compositionality, and logics in sentences. In order to include the above more formally defined properties, additional data augmentation (\citealp{andreas2020good,akyurek2020learning}) or architectural \cite{akyurek2021lexicon} techniques may be needed.

\paragraph{Dense retrieval models and evaluations.}
Another downstream task relevant to sentence embedding is dense retrieval. Given sentences or documents, dense retrieval task aims to find the most relevant pairs within a corpus (\citealt{wang2021GPL, wang2021tsdae, thakur2021augmented, izacard2021towards,liu2022retromae}). Due to the way retrieval tasks are defined, models are trained with different data (Book Corpus, English Wikipedia (\citealt{gao2021condenser,zhu2015aligning})) and the objective encourages high scores given positive pairs, while (our) sentence embedding objective focuses on differentiating sentence semantics. Due to this subtle difference and project scope, we do not evaluate directly on retrieval tasks, and focus on comparing to previous works in the sentence embedding space. 

\paragraph{Choice of backbone models.}
We recognize that there have been many pretrained language models that have out-performed BERT. We used BERT and RoBERTa to make our evaluation comparable to previous works. Finetuning on additional models could lead to insights in trade-offs between pretraining objectives, data size and contrastive finetuning. We leave that for future studies.

\paragraph{Training data size and contrastive finetuning.}
Our method is able to produce SOTA results given a small fine-tuning dataset. However, we were unable to beat other methods that were trained/fine-tuned on much larger datasets.
It is important to note, that \citet{giorgi2021declutr} reported RoBERTa\textsubscript{base} to score 87.31 on average transfer results. This indicates that finetuning RoBERTa with contrastive objective on wiki1m \textbf{reduces} the transfer performance (for SimCSE, DiffCSE, and AugCSE). One potential explanation for such behavior is that RoBERTa is trained on a much larger dataset with carefully designed next-sentence prediction objective, and has learned a robust sentence embedding already (given cpt-text-S was finetuned solely based on signals between neighboring sentences).


\paragraph{Language in concern}
During our study we limited our exploration to English only for better comparison to previous works. However, NLAugmentor does provide many augmentations that are focused on non-English, or multiple languages (which we filtered out for the scope of our project and training dataset). Nonetheless, our results could be extended to improving multi-lingual sentence embedding representations given the right training data and augmentation that can improve downstream multilingual tasks such as multilingual semantic textual similarity \cite{cer2017semeval}, parallel corpus mining, a similar task to dense retrieval tasks in multilingual corpora (\citealt{zweigenbaum-etal-2017-overview,zweigenbaum2018overview,10.1162/tacl_a_00288,reimers2020making,jones2021majority,feng2022language}), machine translation (MT) and MT Quality Estimate (MTQE) that predicts the quality of the output provided by an MT system at test time when no gold-standard human translation is available \cite{fomicheva2020unsupervised,kocyigit2022better}. In fact, one of the main domains in which we believe our methods could come into use is in low-resource languages. Previous works have typically used backtranslation \cite{sennrich-etal-2016-improving} and comparable corpora (recent works such as \citealt{rasooli2021wikily} and \citealt{kuwanto2103isidora} that also uses code-switch data pre-train their MT encoder) to augment training data in low resource languages MT. In addition, in these settings we can incorporate augmentations that are linguistically rooted (created by language experts) or multi-lingual in nature, to improve neural representations of languages that are not as available as English. 


\section*{Acknowledgements}
We thank Dima Krotov and Yoon Kim for the early ideations of the project and Yoon Kim for helpful feedback before the final submission. We thank Dina Bashkirova, Andrea Burns, and Feyza Akyürek for guidance on experimentation methods. We thank Boston University for providing the computational resource and all anonymous reviewers for their insightful comments in the reviewing process. Zilu is supported by Lu Lingzi Scholarship from Boston University. This work is also supported in part by the U.S. NSF grant 1838193 and DARPA HR001118S0044 (the LwLL program). The U.S. Government is authorized to reproduce and distribute reprints for Governmental purposes. The views and conclusions contained in this publication are those of the authors and should not be interpreted as representing official policies or endorsements of DARPA and the U.S. Government.

\bibliography{anthology,custom}
\bibliographystyle{acl_natbib}
\clearpage
\appendix

\section{Appendix}
\label{sec:appendix}

\subsection{Ethics Statement}

To our best knowledge, there is no outstanding ethical issue with our method of approach other than including potentially problematic augmentations (stereotype-reaffirming, toxic, etc) into the augmentation set. In fact, we believe one of the main advantage of our methods over previous methods is we can use rule-based augmentations to explicitly control for the type of invariances we want to instill within the sentence embedding, as opposed to propagating bias, stereotypes, and toxicity that exist in natural text and pre-trained LMs. NL-Augmenter includes many rule-based augmentations that tackle exactly such biases against country of origin, gender, geolocation, linguistic patterns, etc. 

When considering computing resources and environmental impact, rule-based methods are much cheaper and more accessible to run, making our method a much more desirable approach for low-resource compute settings.

\subsection{All Augmentations Descriptions in Experiments}
\label{apx: aug-desc}
In this section, we \textbf{word-by-word copy over} the descriptions of each of the augmentations we have mentioned in our paper from NL-Augmenter \cite{dhole2021nl}, unless otherwise \textbf{noted}.

\paragraph{SentenceAdjectivesAntonymsSwitch} \label{apx: SentenceAdjectivesAntonymsSwitch}
This transformation switches English adjectives in a sentence with their WordNet \cite{miller1998wordnet} antonyms to generate new sentences with possibly different meanings and can be useful for tasks like Paraphrase Detection, Paraphrase Generation, Semantic Similarity, and Recognizing Textual Entailment.

\underline{Example}: Amanda’s mother was very
\textcolor{red}{beautiful} $\rightarrow$ \textcolor{blue}{ugly} .

\paragraph{SentenceAuxiliaryNegationRemoval} \label{apx: SentenceAuxiliaryNegationRemoval}
This is a low-coverage transformation which targets
sentences that contain negations. It removes negations in English auxiliaries and attempts to generate new sentences with the opposite meaning.

\underline{Example}: Ujjal Dev Dosanjh was \textcolor{red}{not} $\rightarrow$ Ujjal
Dev Dosanjh was the 1st Premier of
British Columbia from 1871 to 1872.

\paragraph{ReplaceHypernyms / ReplaceHyponyms} \label{apx: hypernym/hyponym}
This transformation replaces common nouns with other related words that are either hyponyms or hypernyms. Hyponyms of a word are more specific in meaning (such as a sub-class of the word), eg: ’spoon’ is a hyponym of ’cutlery’. Hypernyms are related words with a broader meaning (such as a generic category /super-class of the word), eg: ’colour’ is a hypernym of ’red’. Not every word will have a hypernym or hyponym.

\paragraph{SentenceSubjectObjectSwitch} \label{apx: SentenceSubjectObjectSwitch}
This transformation switches the subject and object of English sentences to generate new sentences with a very high surface similarity but very different meaning. This can be used, for example, for augmenting data for models that assess semantic similarity

\paragraph{CityNamesTransformation} \label{apx: CityNamesTransformation}
This transformation replaces instances of populous and well-known cities in Spanish and English sentences with instances of less populous and less well-known cities to help reveal demographic biases (Mishra et al., 2020) prevelant in named entity recognition models. The choice of cities have been taken from the World Cities Dataset. \footnote{https://www.kaggle.com/datasets/juanmah/world-cities}

\paragraph{AntonymSubstitute} \label{apx: AntonymSubstitute}
This transformation introduces semantic diversity by replacing an even number of adjective/adverb in a given text. We assume that an even number of antonyms transforms will revert back sentence semantics; however, an odd number of transforms will revert the semantics. Thus, our transform only applies to the sentence that has an even number of revertible adjectives or adverbs.We called this mechanism double negation.

\underline{Example}: Steve is \textcolor{red}{able} $\rightarrow$ \textcolor{blue}{unable} to recommend movies that depicts the lives of \textcolor{red}{beautiful} $\rightarrow$ \textcolor{blue}{ugly} minds.

Note: To increase perturbation rate, and since we discovered that negations in semantics do not change sentence embeddings as much, we modified the original augmentations behavior by changing only odd number of antonyms. Hence, this augmentation changed from "Highly meaning preserving" to "Meaning Alteration". However, after we found out it was very similar to SentenceAjectivesAntonymsSwitch, we did not include it in main experiments for overlapping augmentation.

\paragraph{ColorTransformation} \label{apx: ColorTransformation}
This transformation augments the input sentence by randomly replacing mentioned colors with different ones from the 147 extended color keywords specified by the World Wide Web Consortium (W3C). Some of the colors include “dark sea green”, “misty rose”, “burly wood”.

\underline{Example}: Tom bought 3 apples, 1 \textcolor{red}{orange} $\rightarrow$
\textcolor{blue}{misty rose} , and 4 bananas and paid
\$10.

\paragraph{Summarization} \label{apx: Summarization}
This transformation compresses English sentences by extracting subjects, verbs, and objects of the sentence. It also retains any negations. For example, “Stillwater is not a 2010 American liveaction/animated dark fantasy adventure film” turns into “Stillwater !is film”. \citep{zhang2021smat} used a similar idea to this transformation.

\paragraph{DiverseParaphrase} \label{apx: DiverseParaphrase}
This transformation generates multiple paraphrases of a sentence by employing 4 candidate selection methods on top of a base set of backtranslation models. 1) DiPS \cite{kumar2019submodular} 2) Diverse Beam Search \cite{vijayakumar2018diverse} 3) Beam Search \cite{wiseman2016sequence} 4) Random. Unlike beam search which generally focusses on the top-k candidates, DiPS introduces a novel formulation of using submodular optimisation to focus on generating more diverse paraphrases and has been proven to be an effective data augmenter for tasks like intent recognition and paraphrase detection \cite{kumar2019submodular}. Diverse Beam Search attempts to generate diverse sequences by employing a diversity promoting alternative to the classical beam search \cite{wiseman2016sequence}.

\paragraph{SentenceReordering} \label{apx: SentenceReordering}
This perturbation adds noise to all types of text sources (paragraph, document, etc.) by randomly shuffling the order of sentences in the input text \cite{lewis2020bart}. Sentences are first partially decontextualized by resolving coreference \cite{lee2018higher}. This transformation is limited to input text that has more than one sentence. There are still cases where coreference can not be enough for decontextualization. For example, there could be occurences of ellipsis as demonstrated by \citep{gangal2021nareor} or events could be mentioned in a narrative style which makes it difficult to perform re-ordering or shuffling \cite{kovcisky2018narrativeqa} while keeping the context of the discourse intact.

\paragraph{TenseTransformation} \label{apx: TenseTransformation} 

This transformation converts English sentences from one tense to the other, for example simple present to simple past. This transformation was introduced by \cite{logeswaran2018content}.

\paragraph{RandomDeletion} \label{apx: RandomDeletion} 
This augmentation randomly deletes a proportion of the words \cite{wei2019eda} and was added by us into the library of augmentations. Implementation uses nlpAug \cite{ma2019nlpaug}.

\paragraph{RandomCrop} \label{apx: RandomCrop} 
This augmentation randomly deletes a continuous span of words and was added by us into the library of augmentations. Implementation uses nlpAug \cite{ma2019nlpaug}.

\paragraph{RandomSwap} \label{apx: RandomSwap} 
This augmentation randomly swaps a proportion of the words and was added by us into the library of augmentations. Implementation uses nlpAug \cite{ma2019nlpaug}.

\paragraph{RandomWordAugmentation} \label{apx: RandomWordAugmentation} 
This augmentation transforms input by uniformly randomly select an augmentation from RandomDeletion, RandomCrop, and RrandomSwap. Implementation uses nlpAug \cite{ma2019nlpaug}.

\paragraph{RandomWordEmbAugmentation} \label{apx: RandomWordEmbAugmentation}

This augmentation substitute words with similar words defined by Glove embedding \cite{pennington2014glove}. Implementation uses nlpAug \cite{ma2019nlpaug}.

\paragraph{RandomContextualWordAugmentation} \label{apx: RandomContextualWordAugmentation}
This augmentation randomly masks and fills words with pretrained BERT models. Similar ideas are often used in adversarial word embedding literature \cite{morris2020textattack}. Implementation uses nlpAug \cite{ma2019nlpaug}.

\paragraph{YodaPerturbation} \label{apx: YodaPerturbation}
This perturbation modifies sentences to flip the clauses such that it reads like "Yoda Speak". For example, "Much to learn, you still have". This form of construction is sometimes called "XSV", where "the “X” being a stand-in for whatever chunk of the sentence goes with the verb", and appears very rarely in English normally. The rarity of this construction in ordinary language makes it particularly well suited for NL augmentation and serves as a relatively easy but potentially powerful test of robustness.

\paragraph{ContractionExpansions} \label{apx: ContractionExpansions}
This perturbation substitutes the text with popular expansions and contractions, e.g., “I’m” is changed to “I am”and vice versa. The list of commonly used contractions \& expansions and the implementation of perturbation has been taken from Checklist \cite{ribeiro2020beyond}.

\underline{Example:} He often does \textcolor{red}{n’t} $\rightarrow$ \textcolor{blue}{not} come to
school.

\paragraph{DiscourseMarkerSubstitution} \label{apx: DiscourseMarkerSubstitution}

This perturbation replaces a discourse marker in a sentence by a semantically equivalent marker. Previous work has identified discourse markers that have low ambiguity \cite{pitler2008easily}. This transformation uses the corpus analysis on PDTB 2.0 \cite{prasad2008penn} to identify discourse markers that are associated with a discourse relation with a chance of at least 0.5. Then, a marker is replaced with a different marker that is associated to the same semantic class.

\underline{Example}: It has plunged 13\% \textcolor{red}{since} $\rightarrow$
\textcolor{blue}{inasmuch} as July to around 26 cents a pound. A year ago ethylene sold for
33 cents

\paragraph{Casual2Formal} \label{apx: Casual2Formal}

This transformation transfers the style of text from
formal to informal and vice versa. It uses the implementation of Styleformer\footnote{https://github.com/PrithivirajDamodaran/Styleformer}.

\underline{Example}: What you \textcolor{red}{upto} $\rightarrow$ \textcolor{blue}{currently doing} ?

\paragraph{GenderSwap} \label{apx: GenderSwap}
This transformation introduces gender diversity to the given data. If used as data augmentation for training, the transformation might mitigate gender bias, as shown in \citep{dinan2020queens}. It also might be used to create a gender-balanced evaluation dataset to expose the gender bias of pre-trained models. This transformation performs lexical substitution of the opposite gender. The list of gender pairs (shepherd <–> shepherdess) is taken from \citep{lu2020gender}. Genderwise names used from \citep{ribeiro2020beyond} are also randomly swapped.

\paragraph{GeoNamesTransformation} \label{apx: GeoNamesTransformation}
This transformation augments the input sentence with information based on location entities (specifically cities and countries) available in the GeoNames database\footnote{http://download.geonames.org/export/dump}. E.g., if a country name is found, the name of the country is appended with information about the country like its capital city, its neighbouring countries, its continent, etc. Some initial ideas of this nature were explored in \citep{puais2019contributions}.

\paragraph{NumericToWord} \label{apx: NumericToWord}

This transformation translates numbers in numeric form to their textual representations. This includes general numbers, long numbers, basic math characters, currency, date, time, phone numbers, etc.

\paragraph{SynonymSubstitution} \label{apx: SynonymSubstitution}

This perturbation randomly substitutes some words in an English text with their WordNet \cite{miller1998wordnet} synonyms \cite{wei2019eda}.

\paragraph{PigLatin} \label{apx: PigLatin}

This transformation translates the original text into pig latin. Pig Latin is a well-known deterministic transformation of English words, and can be viewed as a cipher which can be deciphered by a human with relative ease. The resulting sentences are completely unlike examples typically used in LM training. As such, this augmentation change the input into inputs which are difficult for a LM to interpret, while being relatively easy for a human to interpret.

\paragraph{PhonemeSubstitution} \label{apx: PhonemeSubstitution}
This transformation adds noise to a sentence by randomly converting words to their phonemes.This transformation adds noise to a sentence by randomly converting words to their phonemes. Grapheme-to-phoneme substitution is useful in NLP systems operating on speech. An example of grapheme to phoneme substitution is “permit” $\rightarrow$ P ER0 M IH1 T’.

\paragraph{VisualAttackLetter} \label{apx: VisualAttackLetter}

This perturbation replaces letters with visually similar, but different, letters. Every letter was embedded into 576-dimensions. The nearest neighbors
are obtained through cosine distance. To obtain
the embeddings the letter was resized into a 24x24
image, then flattened and scaled. This follows the
Image Based Character Embedding (ICES) \cite{eger2019text}.
The top neighbors from each letter are chosen.
Some were removed by judgment (e.g. the nearest neighbors for ’v’ are many variations of the
letter ’y’) which did not qualify from the image
embedding \cite{eger2019text}.

\paragraph{BackTranslation} \label{apx: BackTranslation}

This transformation translates a given English sentence into German and back to English.This transformation acts like a light paraphraser. Multiple
variations can be easily created via changing parameters like the language as well as the translation
models which are available in plenty. Backtranslation has been quite popular now and has been a
quick way to augment examples (\citealt{li2019improving},
; \citealt{sugiyama2019data}).

\paragraph{MultilingualBackTranslation} \label{apx: MultilingualBackTranslation}

This transformation translates a given sentence
from a given language into a pivot language and
then back to the original language. This transformation is a simple paraphraser that works on 100 different languages. Back Translation has been quite
popular now and has been a quick way to augment
(\citealt{li2019improving}; \citealt{sugiyama2019data}; \citealt{fan2021beyond}).

\underline{Example}: Being \textcolor{red}{honest} $\rightarrow$ \textcolor{blue}{Honesty} should
be one of our most important
\textcolor{red}{character traits} $\rightarrow$ \textcolor{blue}{characteristics}

\paragraph{FactiveVerbTransformation} \label{apx: FactiveVerbTransformation}

This transformation adds noise to all types if text source (sentence, paragraph, etc.) by adding factive verbs based paraphrases \cite{alvin2012annotating}
\underline{Example}: Peter published a research paper $\rightarrow$ Peter \textcolor{blue}{acknowledged that} he published a research paper.

\subsection{Narrowing down augmentations}
\label{apx: narrow-down}
we first filter for single sentence operations for unsupervised settings. We then remove augmentations that do not represent typical text distributions (\hyperref[apx: PigLatin]{PigLatin}), or perturb based on audio (\hyperref[apx: PhonemeSubstitution]{PhonemeSubstitution}) or visual (\hyperref[apx: VisualAttackLetter]{VisualAttackLetter}) similarities. Since semantic similarities between augmented and original sentence is important to our objective, we categorize all augmentations according to meaning preservation label provided by NL-Augmenter: \textbf{highly meaning preserving}, \textbf{possible meaning alteration}, and \textbf{meaning alteration}. Given not all augmentations were labeled, we manually label missing augmentations. Lastly, we filter out similar methods and only keep one from every type of augmentation (\hyperref[apx: MultilingualBackTranslation]{MultilingualBackTranslation}, \hyperref[apx: BackTranslation]{BackTranslation}, etc.), and keep only augmentations that have relatively high perturbation rates (> 0.2). We then manually look through augmentation examples to filter out augmentations that produce repetitive artifacts that can be exploited by contrastive learning scheme (\hyperref[apx: FactiveVerbTransformation]{FactiveVerbTransformation}).

\subsection{Code for Gradient Reversal Layer}
\label{apx: grl-code}

\begin{lstlisting}[language=Python]
from torch.autograd import Function

class GradReverse(Function):

  @staticmethod
  def forward(ctx, x, lambd, **kwargs: None):
    ctx.lambd = lambd
    return x.view_as(x)

  @staticmethod
  def backward(ctx, *grad_output):
    return grad_output[0] * -ctx.lambd, None
\end{lstlisting}
\footnote{Implementation borrowed from https://zhuanlan.zhihu.com/p/263827804}

\subsection{Code for Discrimimnator MLP}
\label{apx: mlp-code}

\begin{lstlisting}[language=Python]
class ProjectionMLP(nn.Module):
  def __init__(self, hidden_size, alpha=1.0):
    super().__init__()
    in_dim = hidden_size
    middle_dim = hidden_size * 2
    out_dim = hidden_size
    self.net = nn.Sequential(
        nn.Dropout(p=0.2),
        nn.Linear(in_dim, middle_dim),
        nn.Tanh(),
        nn.Dropout(p=0.2),
        nn.Linear(middle_dim, out_dim),
        nn.Tanh(),
    )
    self.alpha = alpha
    
  def forward(self, x):
    x = GradReverse.apply(x, self.alpha)
    return self.net(x)
\end{lstlisting}

\subsection{Hyperparameter Selection}
\label{apx: hyp-selection}
For main STS and transfer results, we follow similar search strategy as SimCSE and DiffCSE. For either tasks, we search for best performing dev runs in the hyperparmeter ranges (STS-b dev performance for STS test results; average transfer dev for transfer test results), and use that hyperparaemter set as the best performing set. The hyperparameter search range include: $\lambda \in \{1e-5,5e-5,1e-4,5e-4,1e-3,5e-3,1e-2\}$, learning rate $\in \{5e-6, 7e-6, 1e-5, 2e-5, 3e-5, 5e-5\}$ and batch size is fixed to 128. After obtaining the best hyperparameter for the task, we run the same trial with seed $\in \{1, 11, 42, 68, 421\}$ to obtain standard deviation and average. In the main result, we report maximum of the 5 seeds. In \ref{apx: main-variance}, we report average and variance of 5 trials.
 
For all ablation experiments, we use the best hyperparameter main results (STS and transfer tasks separately), and search with different $\lambda$ only for the best dev results for each ablation trial, and report the dev performances.

\subsection{Best Hyperparameter for Main Results}
\label{apx: hyp-main}
\begin{table}[t]
    \centering
    \begin{tabular}{ccc}
    \toprule
        hyperparameter  & BERT\textsubscript{base}         & RoBERTa\textsubscript{base}\\ \midrule
        $\lambda$         & 5e-3      & 1e-4       \\ 
        learning rate   & 2e-5      & 2e-5       \\ \bottomrule
    \end{tabular}
    \caption{ Best hyperparameters for main STS-B results. 
    \label{tab: hyp-stsb}}
\end{table}

\begin{table}[t]
    \centering
    \begin{tabular}{ccc}
    \toprule
        hyperparameter  & BERT\textsubscript{base}         & RoBERTa\textsubscript{base}\\ \midrule
        $\lambda$         & 1e-4           & 1e-2       \\ 
        learning rate   & 2e-5           & 7e-6       \\ \bottomrule
    \end{tabular}
    \caption{ Best hyperparameter for main SentEval transfer results. 
    \label{tab: hyp-transfer}}
\end{table}

See Table~\ref{tab: hyp-stsb} and ~\ref{tab: hyp-transfer}

\subsection{Main Result Variance}
\label{apx: main-variance}
\begin{table}[t]
    \centering
    \small
    \begin{tabular}{ccc}
    \toprule
        Mode                              & STS-b           & Transfer    \\ \midrule
        SimCSE /w MLM                     & 76.25           & 86.64       \\ 
        DiffCSE                           & 78.49           & 86.86       \\ 
        AugCSE\textsubscript{BERT}      & 77.27 $\pm$ 0.63  & 86.74 $\pm$ 0.29 \\
        AugCSE\textsubscript{RoBERTa}   & 75.54 $\pm$ 1.67  & 86.07 $\pm$ 0.21 \\ \bottomrule
    \end{tabular}
    \caption{ Main results with standard deviation  \label{tab: main-variance}}
\end{table}

See Table ~\ref{tab: main-variance}

\subsection{Reproducibility}
\label{apx: reproducibility}
All of our models are trained and inferenced on a single NVIDIA V100 GPU per trial. Training a single model for one epoch takes from 40 min to 5 hours, depending on the frequency of evaluation.

\subsection{Model Size}

\begin{table}[t]
    \centering
    \begin{tabular}{ccc}
    \toprule
        Model                           & Train & Inference    \\ \midrule 
        AugCSE\textsubscript{BERT}      & 117M  & 110M \\
        AugCSE\textsubscript{RoBERTa}   & 132M  & 125M \\ \bottomrule
    \end{tabular}
    \caption{ Model Sizes in our experiments \label{tab: model-size}}
\end{table}
See Table~\ref{tab: model-size}

\subsection{Augmentation Unification}

\begin{figure}
\begin{center}
  \includegraphics[width=0.23\textwidth]{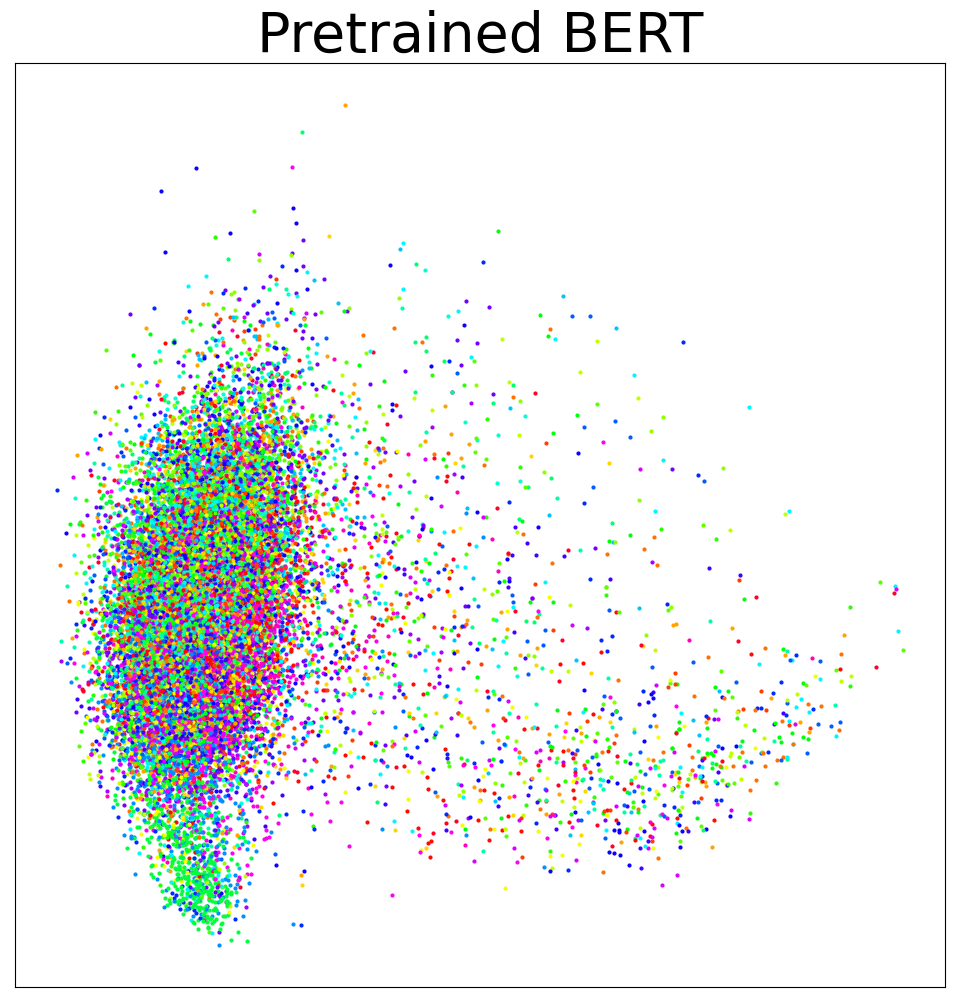}
  \includegraphics[width=0.23\textwidth]{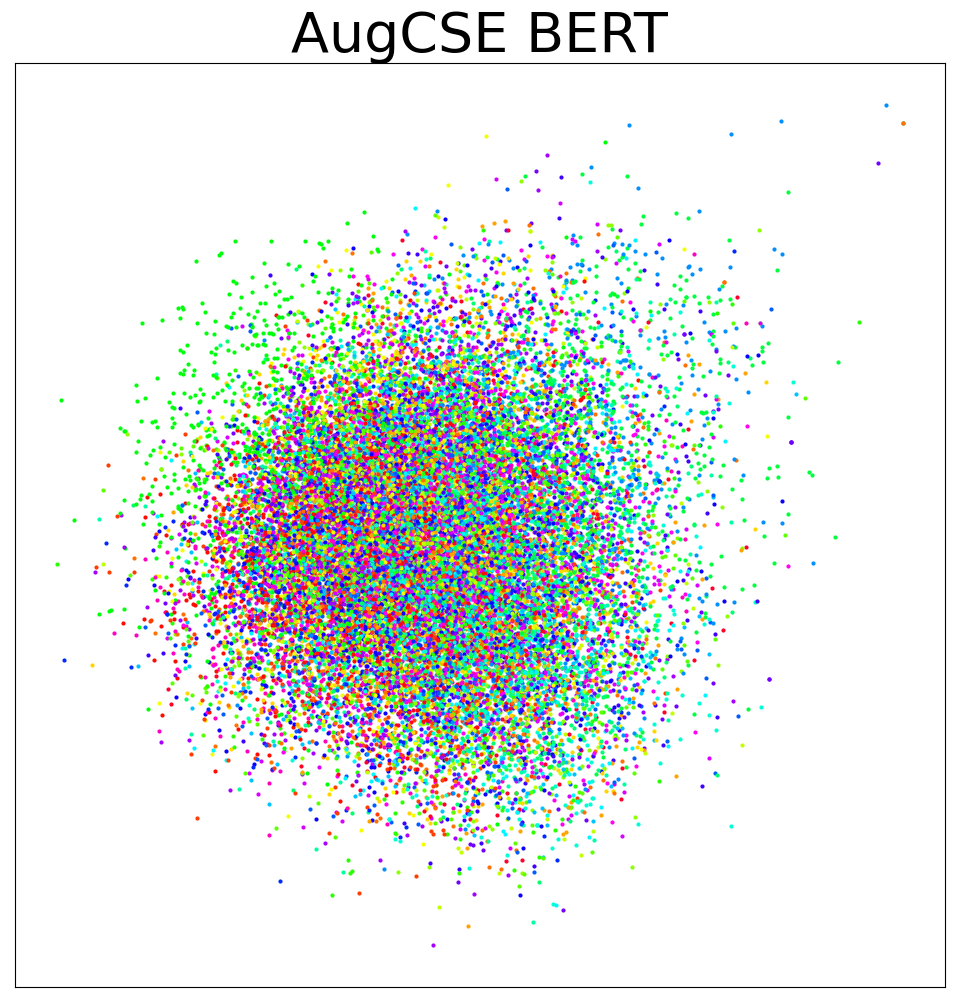}

  \caption{PCA of randomly sampled sentence embeddings from wiki1m dataset with various augmentations (27 augmentations) along with original sentence samples. Color indicates various augmentation types.}
  \label{fig:pca}
\end{center}
\end{figure}

\begin{figure*}
\begin{center}
  \includegraphics[width=0.31\textwidth]{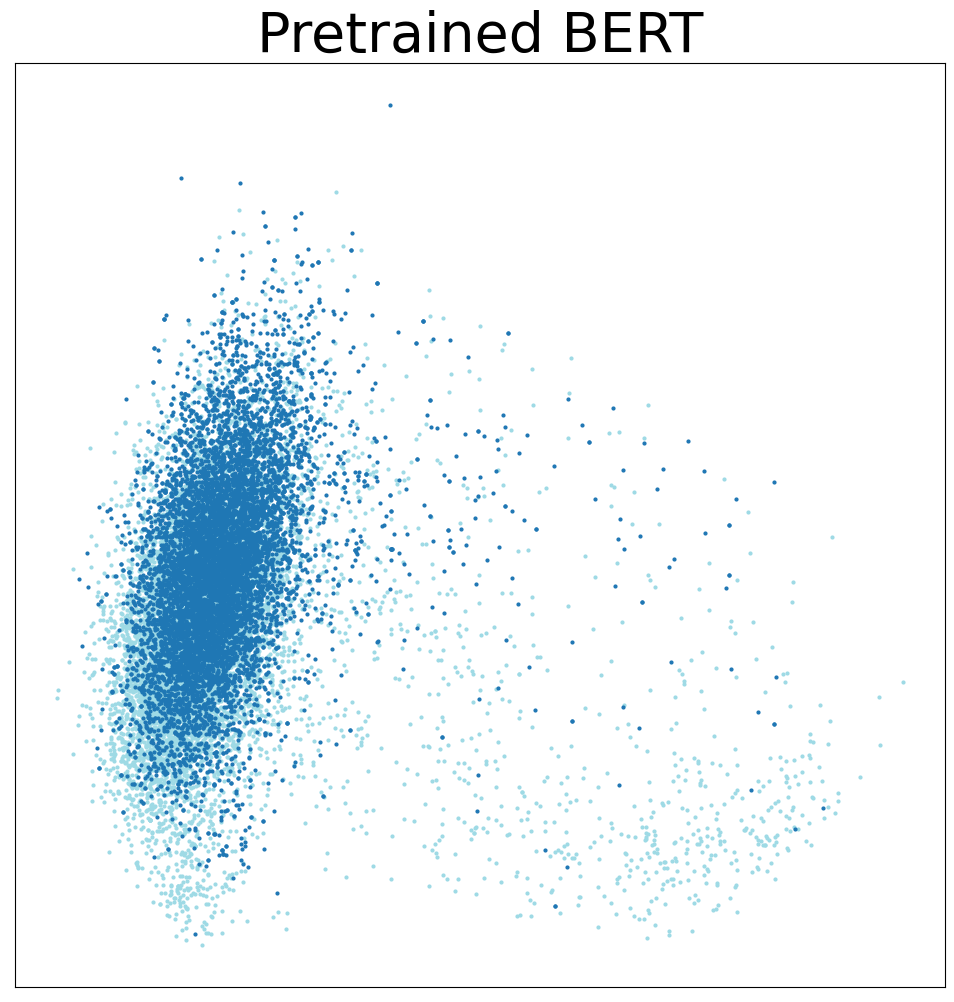}
  \includegraphics[width=0.31\textwidth]{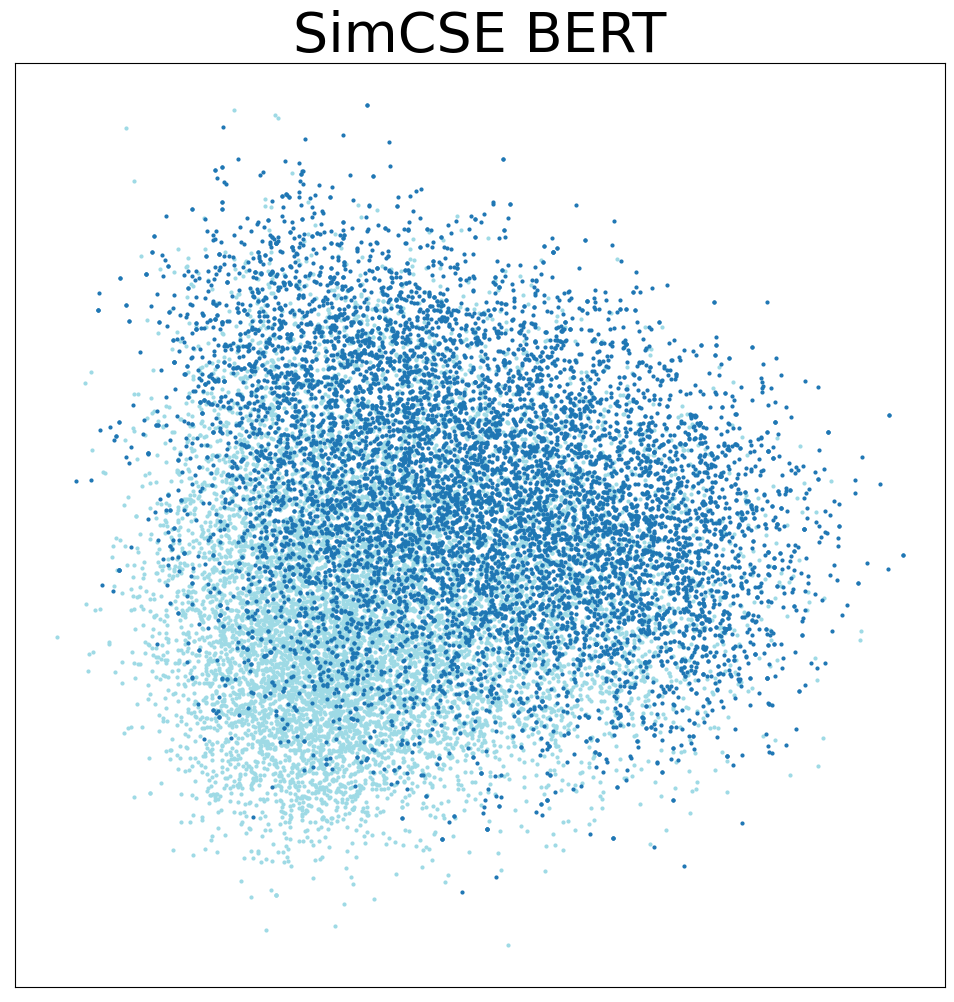}
  \includegraphics[width=0.31\textwidth]{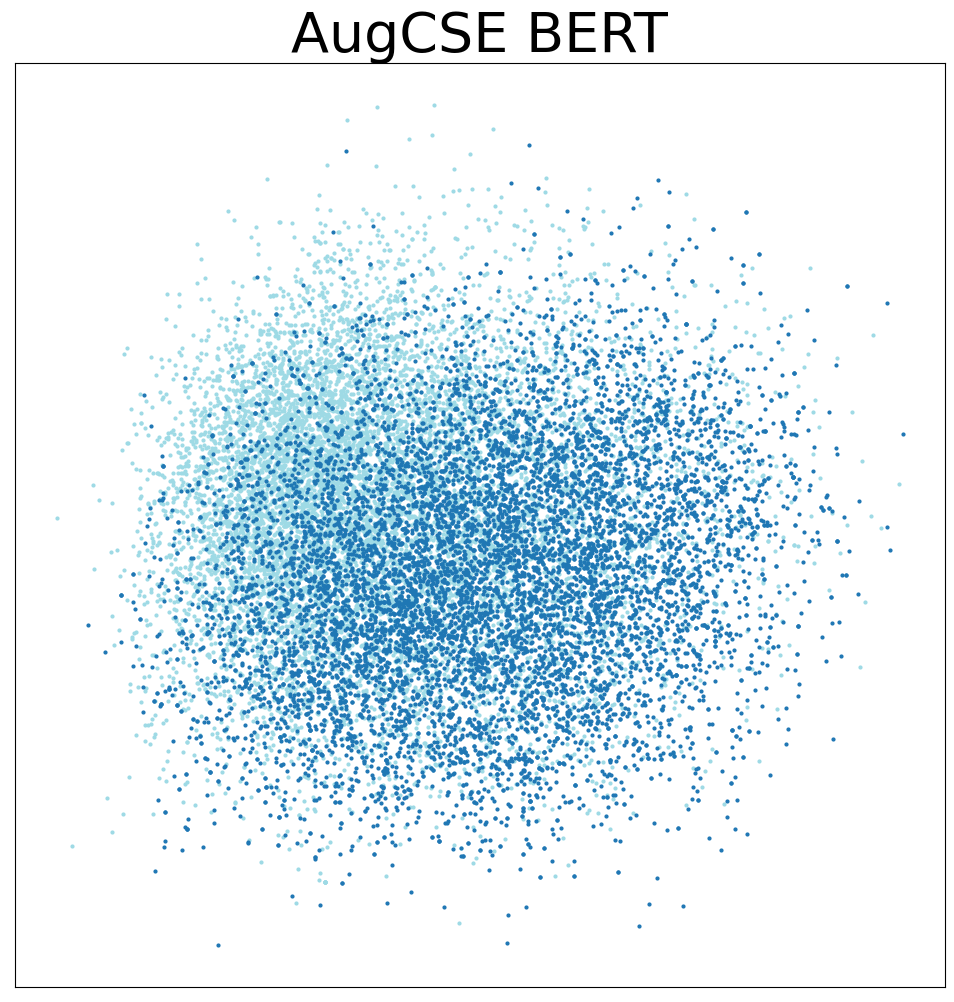}
  \includegraphics[width=0.31\textwidth]{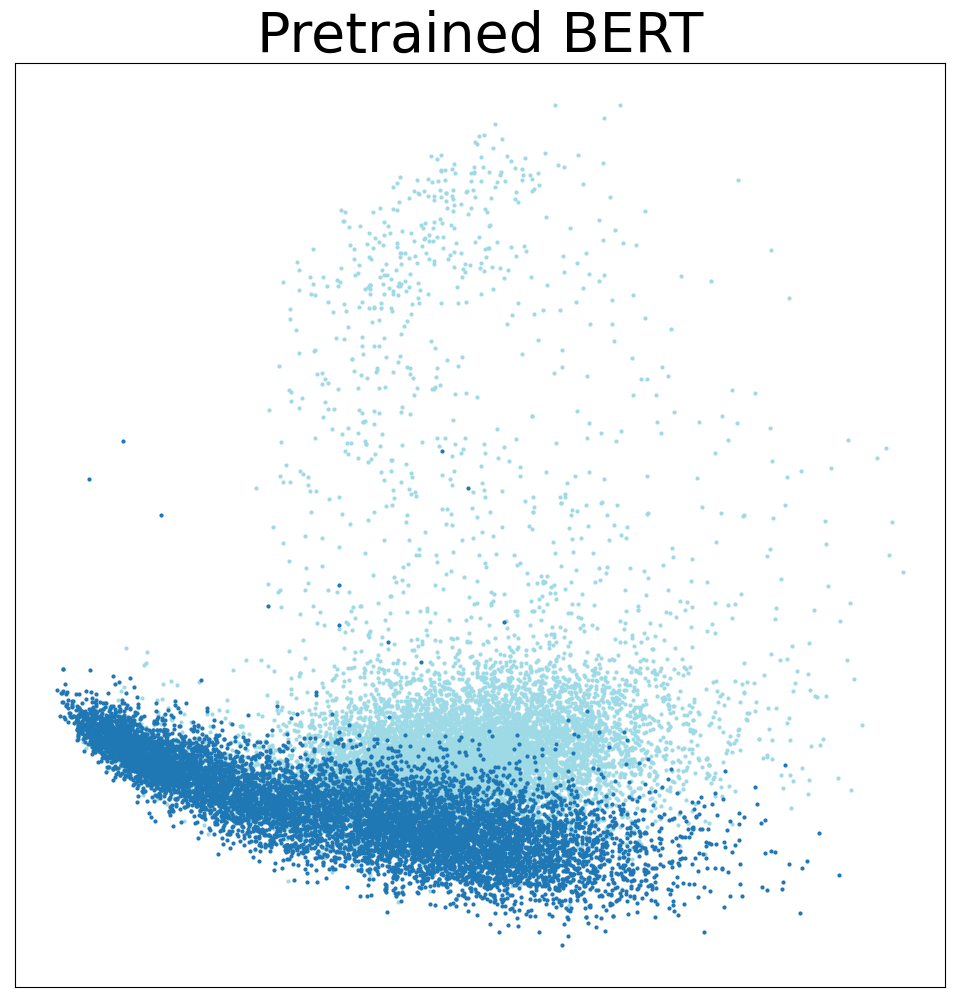}
  \includegraphics[width=0.31\textwidth]{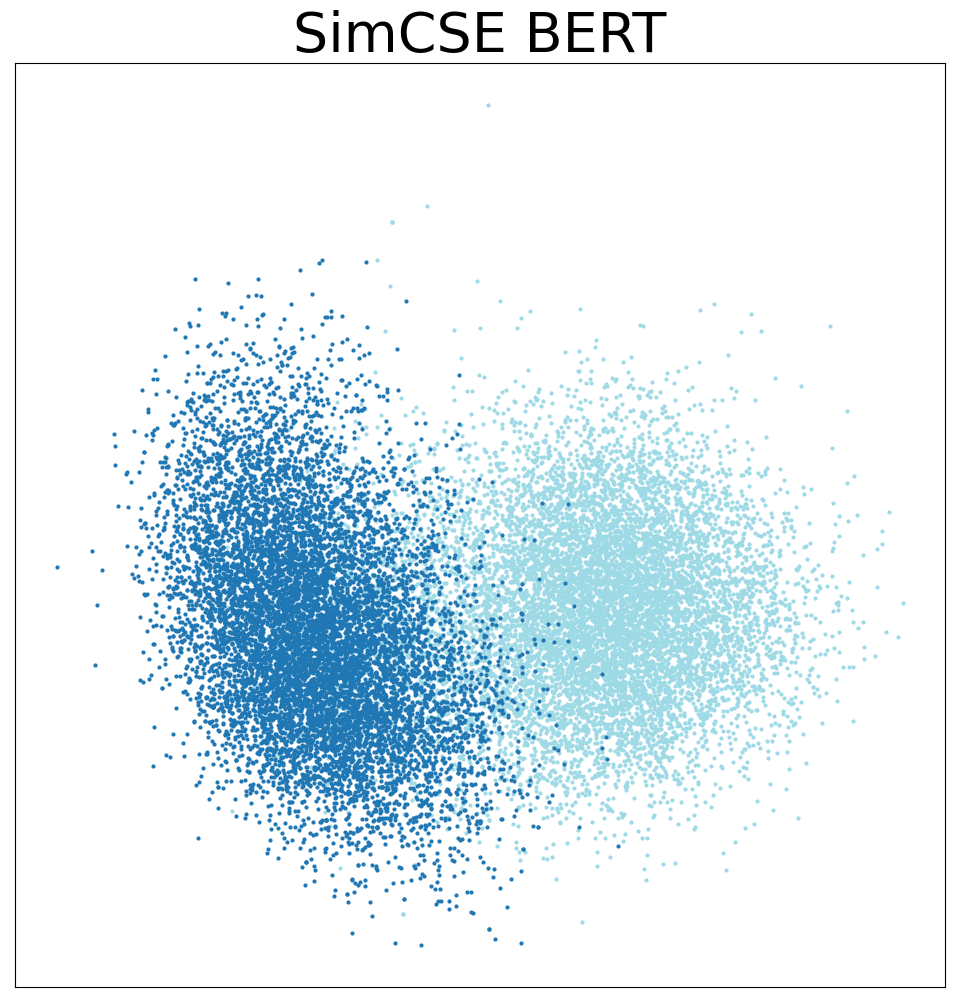}
  \includegraphics[width=0.31\textwidth]{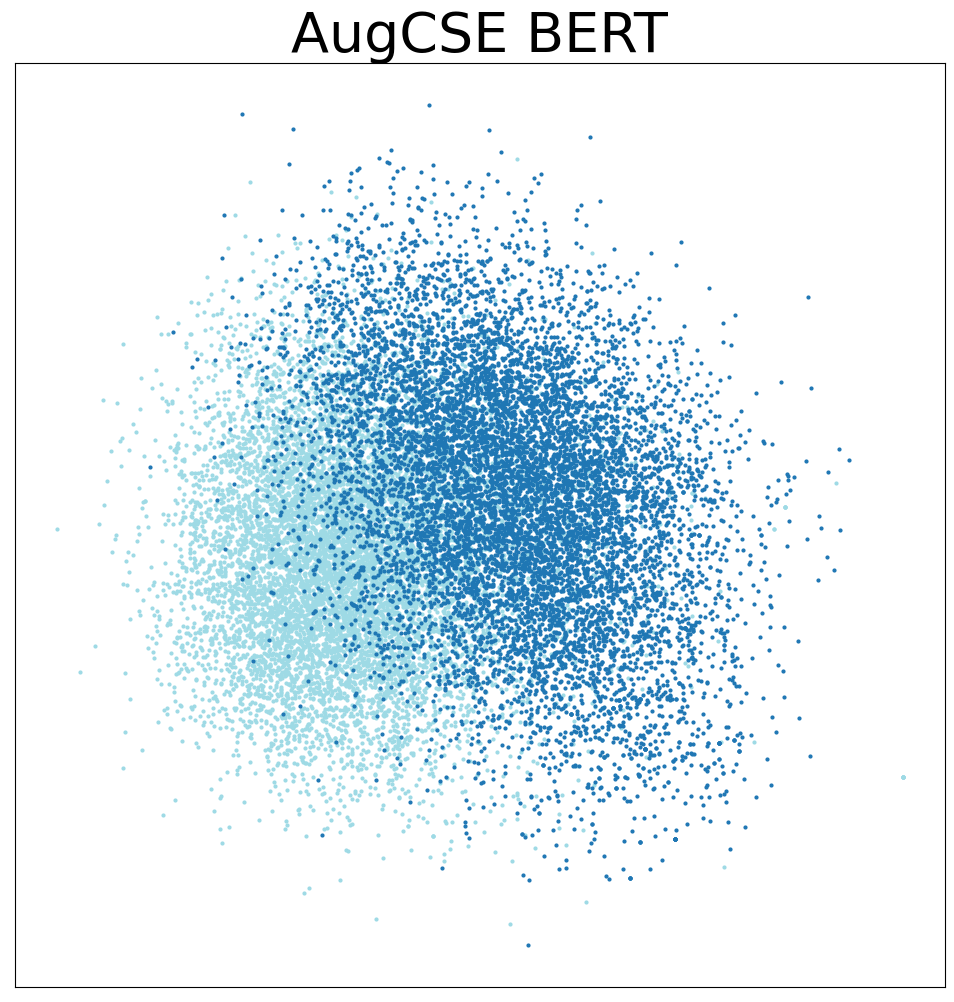}
  \caption{Embedding PCA plot with original sentences and augmented sentences. The augmentation in top row is \hyperref[apx: SentenceAuxiliaryNegationRemoval]{SentenceAuxiliaryNegationRemoval}, and in bottom row is \hyperref[apx: Summarization]{Summarization}}
  \label{fig: single-aug-emb}
\end{center}
\end{figure*}

In Figure~\ref{fig:pca}, we see AugCSE indeed can unify the distribution from different augmentations compare to baseline BERT. In Figure~\ref{fig: single-aug-emb}, we can see that in addition to contrastive objective from SimCSE (and baseline BERT), AugCSE brings distributions of augmentations vs. unperturbed sentences even closer together.

\subsection{Importance of Gradient Reverse Multiplier}
\begin{figure*}
\begin{center}
  \includegraphics[width=0.31\textwidth]{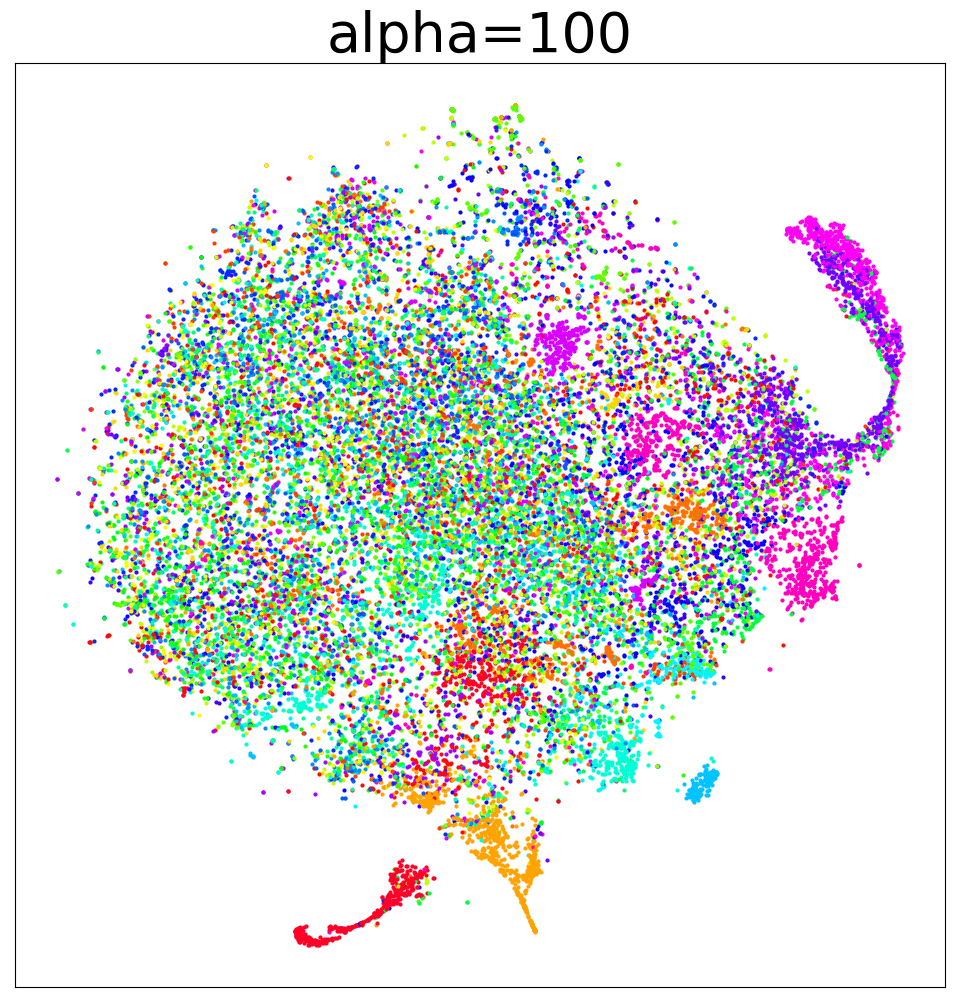}
  \includegraphics[width=0.31\textwidth]{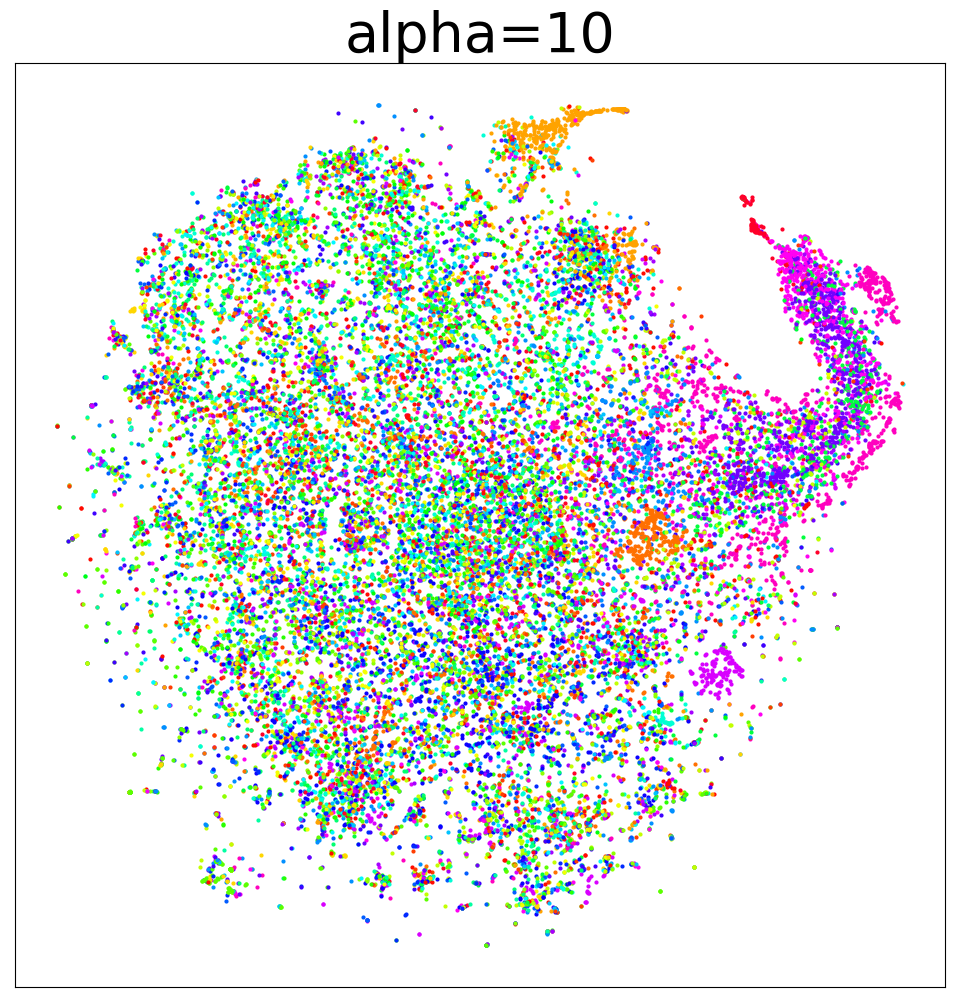}
  \includegraphics[width=0.31\textwidth]{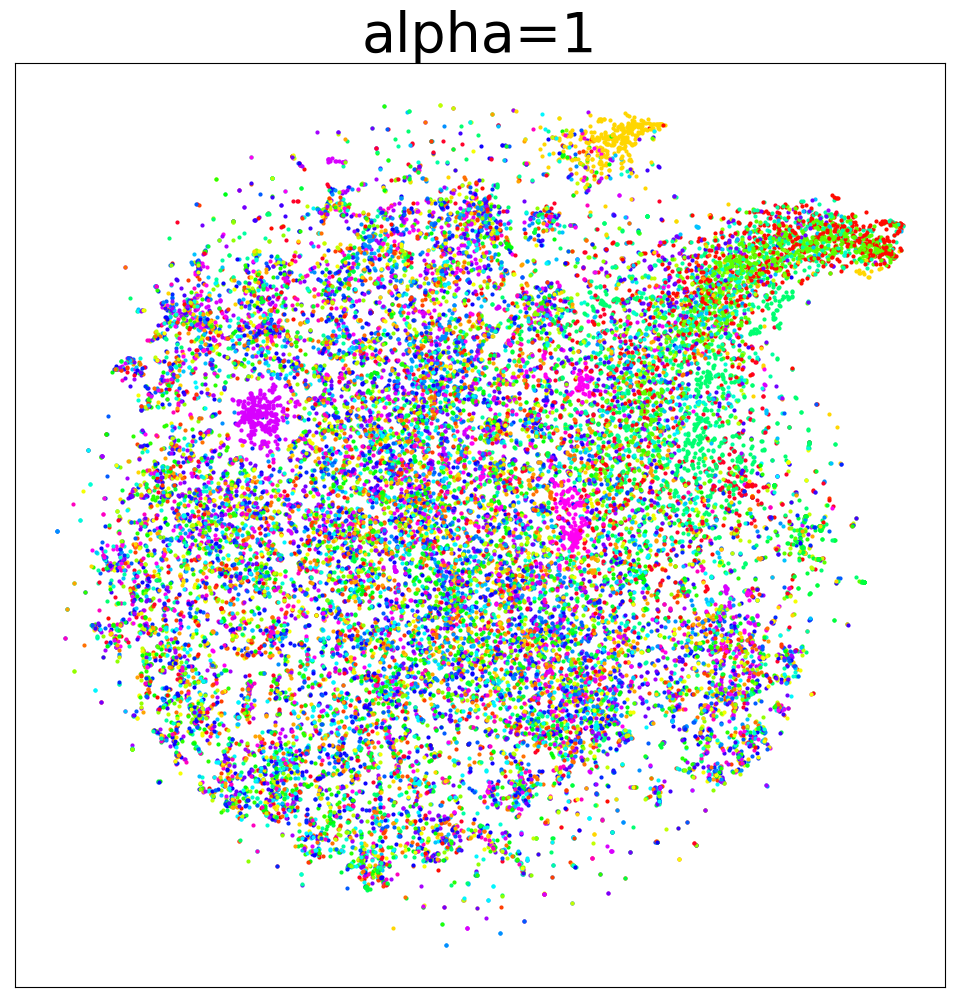}
  \includegraphics[width=0.31\textwidth]{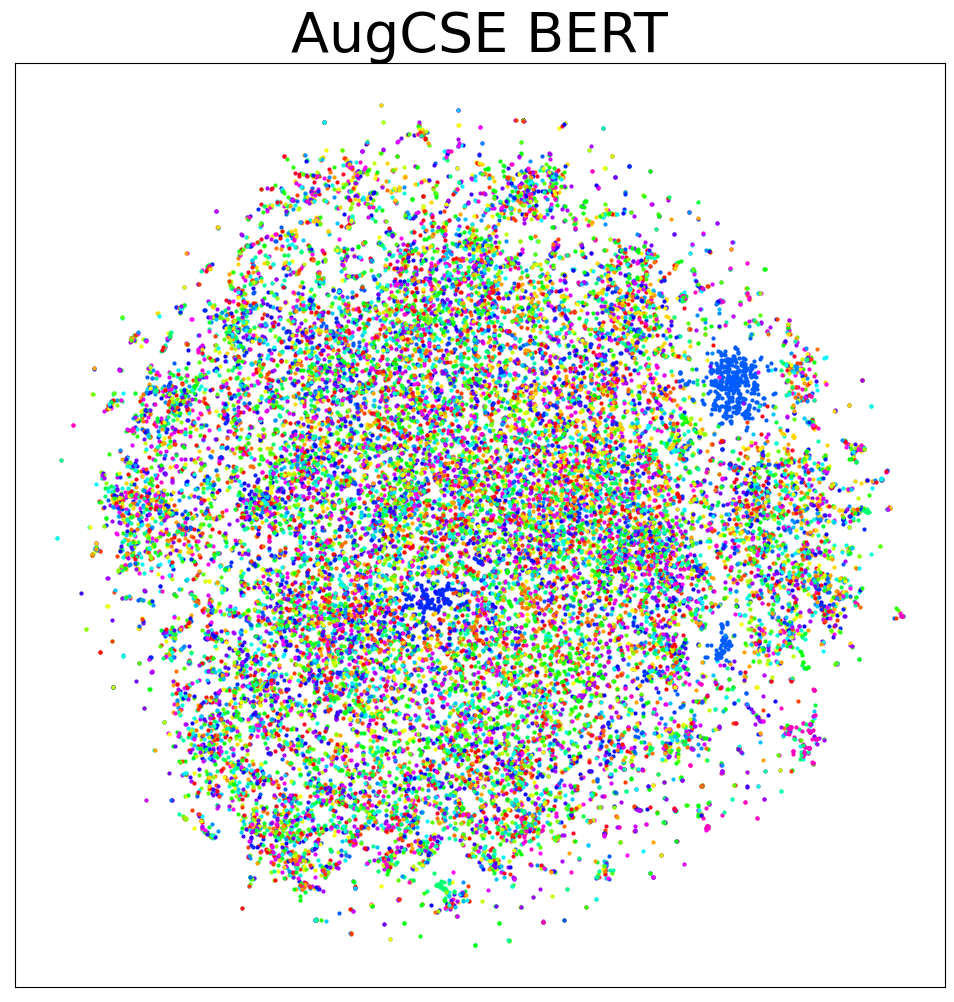}
  \includegraphics[width=0.31\textwidth]{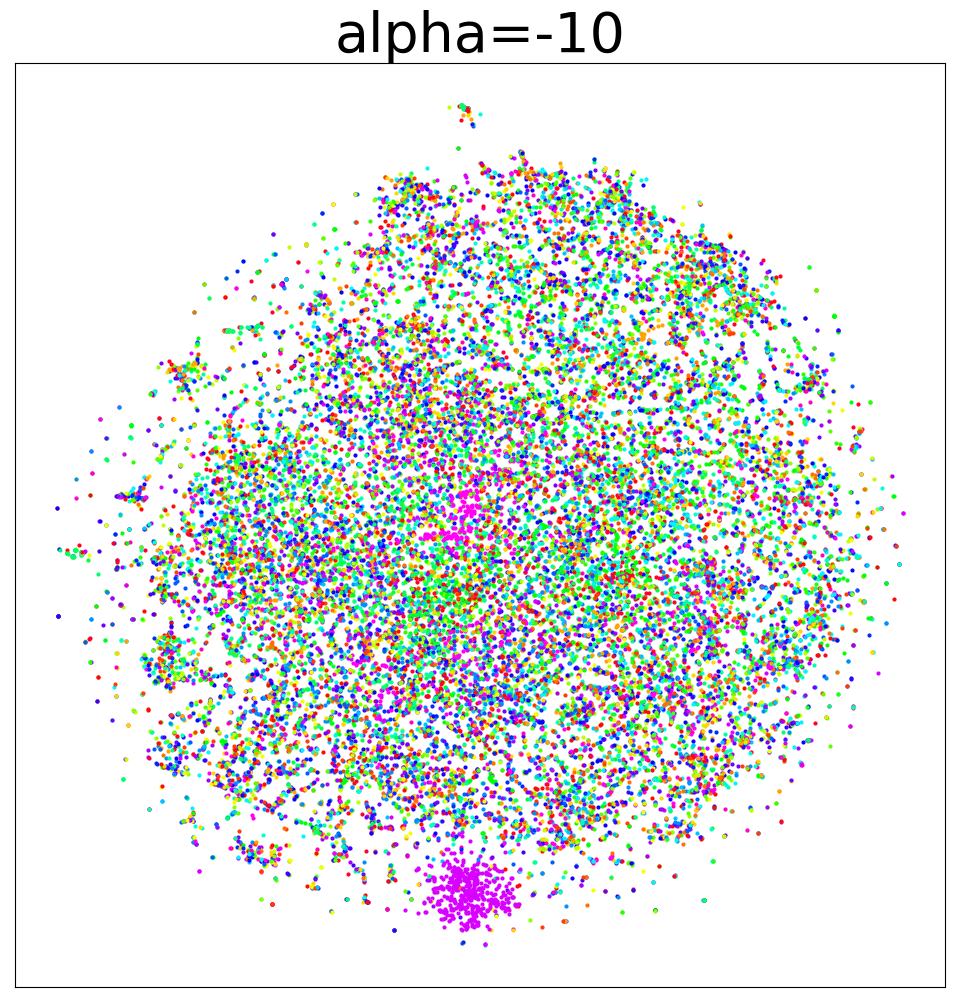}
  \includegraphics[width=0.31\textwidth]{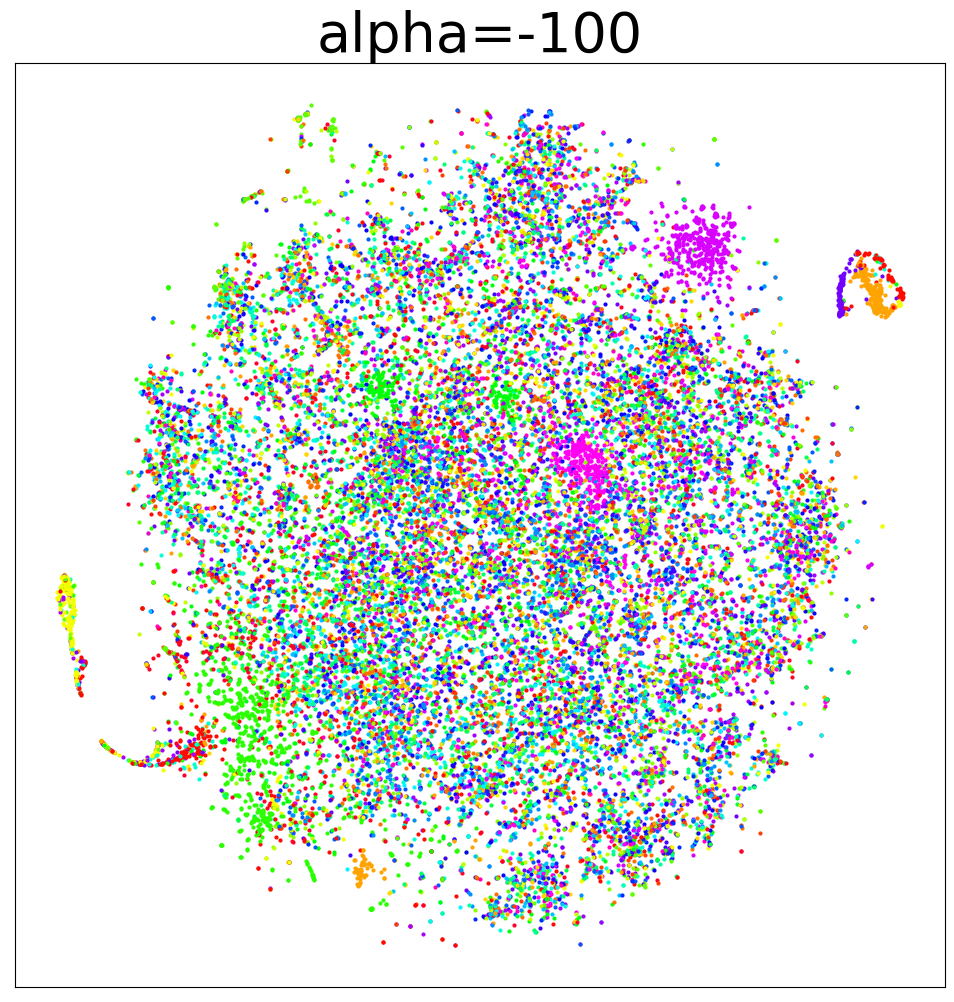}
  \caption{Embedding TSNE plot with different alphas. Colors indicate different augmentation types. Antagonistic discriminators (negative $\alpha$) result in embedding spaces that are more invariant to augmentation types than collaborative discriminators (positive $\alpha$).}
  \label{fig: alpha-emb}
\end{center}
\end{figure*}

\begin{figure*}
\begin{center}
  \includegraphics[width=0.6\textwidth]{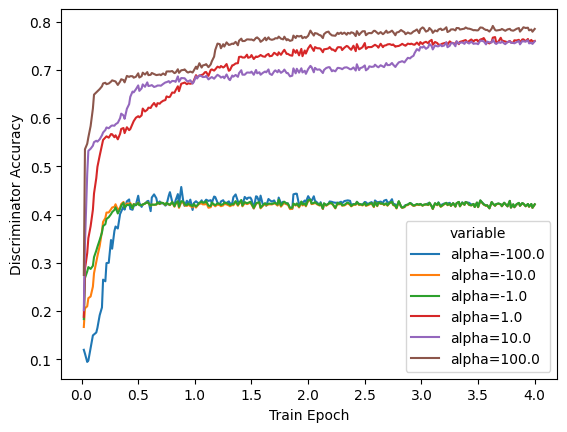}
  \caption{Discriminator accuracy over training with different alpha values.}
  \label{fig: alpha-train}
\end{center}
\end{figure*}

As seen in both training plots (Table~\ref{fig: alpha-train}, Figure~\ref{fig: alpha-emb}), a positive alpha value (collaborative discriminator) results in embeddings that are easily classified by augmentations, whereas negative alpha values (antagonistic discriminator) results in unified embedding that is harder to pick out augmentation type. We use sklearn PCA module for all PCA results, and Multcore-TSNE \footnote{https://github.com/DmitryUlyanov/Multicore-TSNE} for ann TSNE plots.

\subsection{Embedding isomorphism}
\label{apx: isomorphism}

Different augmentations and datasets have been proposed as positive or negative pairs to learn sentence embedding. However, their performance differ drastically, despite many of them were created with the same original purpose, such as paraphrase. In search for what causes the difference in performance, we investigate further in NLI datasets, specifically ANLI \cite{nie2020adversarial}, which was created with the same objective (entailment and contradiction) but with drastically different method. In ANLI, anchor sentences were provided, and entailment and contradictions were crowd-sourced for the purpose of fooling existing models. With such objective, sentences in contradiction and entailment may come from a different distribution as the anchor sentence. 

We trained SimCSE using ANLI data only, and found ANLI-SimCSE to perform much worse than Supervised SimCSE (trained with MNLI and SNLI), even if we sample and adjust for dataset size difference (Table ~\ref{tab:simcse-anli}). 

\begin{table}[t]
    \centering
    \begin{tabular}{ccc}
    \toprule
        Trial                       & STS-b \\ \midrule
        Unsupervised SimCSE         & 81.18 \\ 
        Supervised SimCSE           & 85.64 \\ 
        Supervised SimCSE (Sampled) & 83.82 \\
        ANLI-SimCSE                 & 75.99 \\
        ANLI-SimCSE w/o negatives   & 78.66 \\
        \bottomrule
    \end{tabular}
    \caption{ Ablation experiments removing symmetric loss. All results are reproduced by us.
    \label{tab:simcse-anli}}
\end{table}

To measure some aspect of distributional shift in the embedding space, we used 3 embedding isomorphism measurements: harmonic mean of effective condition numbers \textbf{COND-HM}, singular value gap \textbf{SVG}, and Gromov-Hausdorff distance \textbf{GH} (\citealt{dubossarsky2020secret, jones2021massively}). 

\begin{table}[t]
    \centering
    \begin{tabular}{cccc}
    \toprule
        Trial    & A-E    & A-C     & E-C    \\ \midrule
        \multicolumn{4}{c}{MNLI + SNLI (sample)} \\ \midrule
        COND-HM  & 94.7 & 95.1 & 95.7 \\ 
        SVG      & 0.87 & 0.84 & 0.59 \\ 
        GD       & 0.31 & 0.29 & 0.05 \\ \midrule
        \multicolumn{4}{c}{ANLI} \\ \midrule
        COND-HM  & 96.0 & 95.7 & 91.54 \\ 
        SVG      & 0.86 & 0.82 & 0.29 \\ 
        GD       & 51.7 & 51.3 & 0.02 \\ 
        \bottomrule
    \end{tabular}
    \caption{ Embedding isomorphism distance comparison between MNLI+SNLI to ANLI. A=Anchor, E=Entailment, C=Contradiction
    \label{tab: emb-dist-anli}}
\end{table}

Seen in Table ~\ref{tab: emb-dist-anli}, for ANLI, entailment and contradictions distributions were much more different from anchors than the that for NLI. We believe this difference could be one of the reason using ANLI examples do not work as well as NLI examples. In another word, in ANLI, perhaps because the embedding difference between contradiction and entailment sentences are so much smaller than both to anchor, that the contrasting signals from positives and negatives are conflicting rather than working together. This hypothesis can be confirmed with ANLI-SimCSE w/o negatives performing better than the trial with negatives.

In similar veins, we investigate whether the same measurement could be indicative of augmentation performance. However, were weren't able to find significant correlation. See the next section for more details.

\subsection{Single augmentation performance and embedding distance}
\label{apx: single-aug}

For single augmentation experiments, we remove data points that are not transformed by the augmentation. We find this to work better than leaving some datapoints un-perturbed, which adds noise to the contrastive objective. In addition, we used symmetric contrastive loss similar to CLIP \cite{radford2021learning}. This improves performance because augmentations introduce distributional shifts in the embedding space that benefits from a symmetric regularization.

\begin{figure*}
  \includegraphics[width=\textwidth]{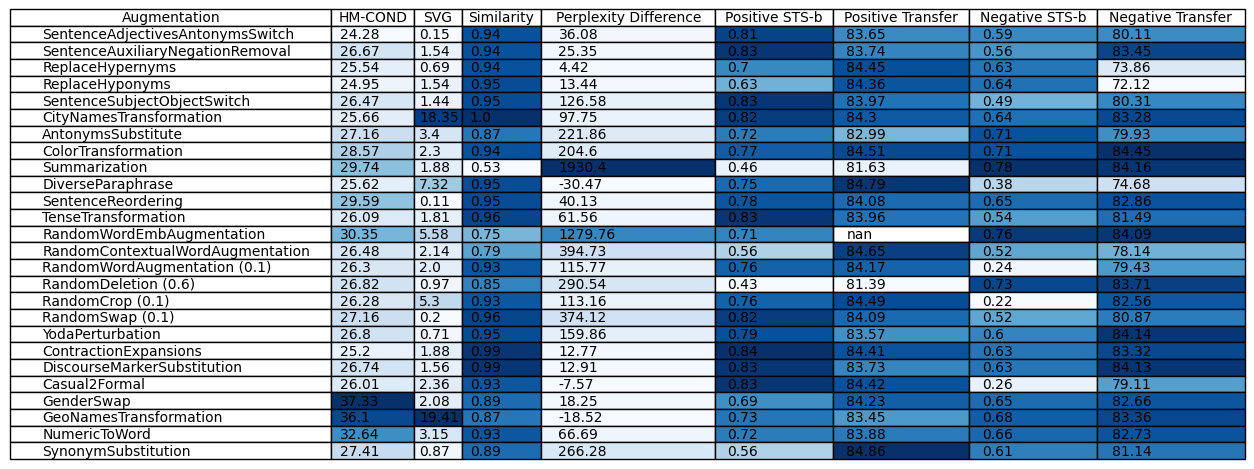}
  \caption{Single augmentation as positive or negative pair in contrastive framework. No discriminator is used. When an augmentation is used as a negative augmentation, the corresponding positive augmentation is the original sentence itself with dropout (SimCSE). The float in parenthesis next to augmentation name indicates the rate of perturbation. \textbf{HM-COND}=harmonic mean of effective condition numbers between augmented and non-augmented sentence embedding samples. \textbf{SVG}=singular value gap between augmented and non-augmented sentence embedding samples. \textbf{Similarity}=cosine similarity of sentence embedding before and after augmentation. \textbf{Perplexity Difference}=perplexity of augmented sentence subtracted by perplexity of original sentence.}
  \label{fig: single-aug-full-table}
\end{figure*}

\begin{figure*}
\begin{center}
  \includegraphics[width=0.7\textwidth]{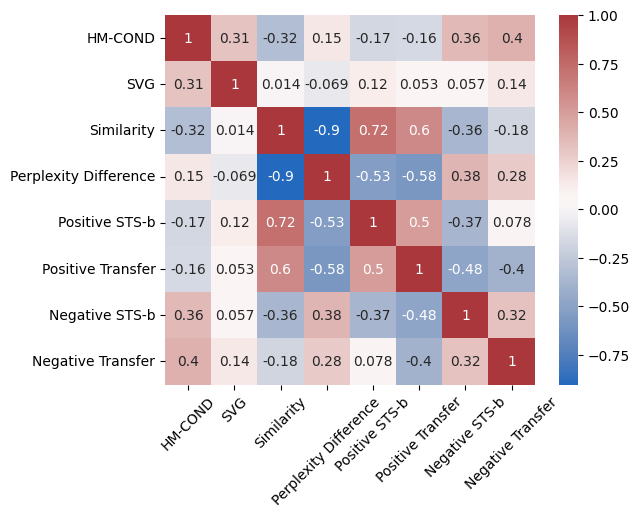}
  \caption{Pearson correlations between columns in Figure~\ref{fig: single-aug-full-table} across all single augmentation trials.}
  \label{fig: single-aug-correlation}
\end{center}
\end{figure*}

In Figure~\ref{fig: single-aug-correlation}, we can observe that similarity and perplexity difference are two measures most correlated feature with respect to all four metrics. Similarity is positively correlated with positive evaluations and perplexity difference is negatively correlated with positive evaluations. Both metrics relation with negative evaluations reverse directions but become much less strongly correlated. This is likely due to the nature of positive and negative augmentation usage in the contrastive objective. The negatives are aggregated along with rest of in-batch examples, lessen the effect. Additionally, the value of negatives is contextually dependent on positives, since the repulsion and attraction of negatives and positives conjointly defines the direction in which anchor embeddings go. HM-COND is also somewhat positively correlated with the with evaluation performance when using augmentation as negatives. It seems to suggest that the more isomorphic the embedding spaces are between augmented vs. original sentences, the better the augmentation is as a negative augmentation.

\subsection{Negations in deep learning}
\label{apx: negation}

As seen in Table~\ref{tab:ablation-simcse-nli}, using contradiction as negatives obtains almost baseline performance, while being semantically entirely opposite. Similarly, in Appendix~\ref{apx: single-aug}, we have also observed that meaning preservation label (Table~\ref{tab: aug-list}) has little indication of whether the augmentation performs well as a single positives. This is a particular interesting phenomenon that requires further study. While a sentence can represent semantically exactly opposite meaning, it is still discussing similar topics, and due to the symmetric nature of cosine similarity, it is difficult to use negation in deep learning. Negative examples do not help as much as in-context learning \cite{wang2022benchmarking} or reinforcement learning rewards \cite{sumers2021learning}, and negative natural language commands lead to exact opposite output from systems \footnote{twitter.com/benjamin\_hilton/status/1520469352008634373}. In toxicity NLP literature, this is related to the phenomenon that superficial textual token meanings are naively combined to yield sentence meaning, without taking to account of deeper structural relationships between entities mentioned \cite{hartvigsen2022toxigen}. In the contrastive learning setting, providing a positive anchor (\textbf{SimCSE} in Table~\ref{tab:ablation-simcse-nli}) helps direct the contrast to a specific direction against the positive examples, yet it is unclear how negatives can be used in other scenarios in deep learning. Such topic could also have interesting implications to "the white bear problem" \cite{wegner2003white}, the phenomenon where "when someone is actively trying not to think of a white bear they may actually be more likely to imagine one." \footnote{en.wikipedia.org/wiki/Ironic\_process\_theory} in psychology, and whether failing to learn from negation in deep learning is a result of in-proper training methods or an indication that deep-learning models are aligned with human psychology, and to solve such problem may require human-centric strategies to deal with such short-comings.

\end{document}